\def\year{2022}\relax
\documentclass[letterpaper]{article}
\usepackage{aaai22} 
\usepackage{times} 
\usepackage{helvet} 
\usepackage{courier} 
\usepackage[hyphens]{url} 
\usepackage{graphicx} 
\usepackage{graphicx}  
\usepackage{natbib}  
\usepackage{caption}  
\DeclareCaptionStyle{ruled}%
{labelfont=normalfont,labelsep=colon,strut=off} 
\usepackage{times}
\usepackage{helvet}
\usepackage{courier}
\usepackage{xcolor}
\usepackage{booktabs}
\usepackage{amsmath,amssymb,amsfonts,amsthm}
\usepackage{enumitem}
\usepackage{algorithm,algorithmic}
\usepackage{subfig}

\newcommand{\argmax}{\operatornamewithlimits{argmax}}
\newcommand{\argmin}{\operatornamewithlimits{argmin}}

\newcommand{\pr}{\mathsf{Pr}}

\newcommand*{\bh}{\mathbf{h}}

\newcommand*{\bx}{\mathbf{x}}

\newcommand*{\bz}{\mathbf{z}}
\newcommand{\kl}[2]{\textnormal{KL}({#1}  \| {#2})}
\newtheorem{thm}{Theorem}

\begin{document}
%
\title{Explaining GNN over Evolving Graphs using Information Flow}
\author{Yazheng Liu$^\ast$, Xi Zhang$^\ast$, Sihong Xie$^\dagger$\\
$^\ast$ BUPT, $^\dagger$ Lehigh University
}
\maketitle
\begin{abstract}
\begin{quote}
Graphs are ubiquitous in many applications, such as social networks, knowledge graphs, smart grids, etc.. Graph neural networks (GNN) are the current state-of-the-art for these applications, and yet
remain obscure to humans.
Explaining the GNN predictions can add transparency.
However, as many graphs are not static but continuously evolving,
explaining changes in predictions between two graph snapshots is different but equally important.
Prior methods only explain static predictions or generate coarse or irrelevant explanations for dynamic predictions.
We define the problem of explaining evolving GNN predictions
and propose an axiomatic attribution method to uniquely decompose the change in a prediction to paths on computation graphs.
The attribution to many paths involving high-degree nodes is still not interpretable,
while simply selecting the top important paths can be suboptimal in approximate the change.
We formulate a novel convex optimization problem to optimally select the paths that explain the prediction evolution.
Theoretically, we prove that the existing method based on Layer-Relevance-Propagation (LRP) is a special case of the proposed algorithm when an empty graph is compared with.
Empirically, on seven graph datasets, with a novel metric designed for evaluating explanations of prediction change, we demonstrate the superiority of the proposed approach over existing methods, including LRP, DeepLIFT, and other path selection methods.
\end{quote}
\end{abstract}

\section{Introduction}
Graph neural networks (GNN) are now the state-of-the-art method for graph representation in many applications, such as molecule property prediction~\cite{tox21}, social network modeling~\cite{kipf2017_iclr}, pose estimation in computer vision~\cite{Yang2021_cvpr}, knowledge graph embedding~\cite{Wang2019_kdd}, and recommendation systems~\cite{Ying2018}.
It is desirable to make the GNN predictions transparent to humans~\cite{gnnexplainer,gnnlrp,Pope2019_cvpr,ren21b,liu2021dig}.
For example,
a user may want to know why a particular recommendation is made to ensure that the recommendation does not breach personal private information (e.g., age and gender);
for sanity check, a prediction of pose must be based on salient configurations of human landmarks rather than backgrounds. 
However, the graph dependencies intertwine a GNN prediction with many nodes several hops away through multiple paths, making the identification of the salient contributors challenging.
To explain GNN predictions on a static graph,
there are local or global explanation approaches~\cite{Yuan2020ExplainabilityIG}.
Given a trained GNN model,
local explanations explain a GNN prediction
by selecting salient subgraphs~\cite{gnnexplainer},
nodes, or edges~\cite{gnnlrp}.
Global methods~\cite{yuan2020xgnn,Vu2020PGMExplainerPG,graphlime} learn simpler surrogate models to approximate the target GNN and generate explaining models or instances.

\begin{figure}[t]
    \centering
    \includegraphics[width=0.37\textwidth]{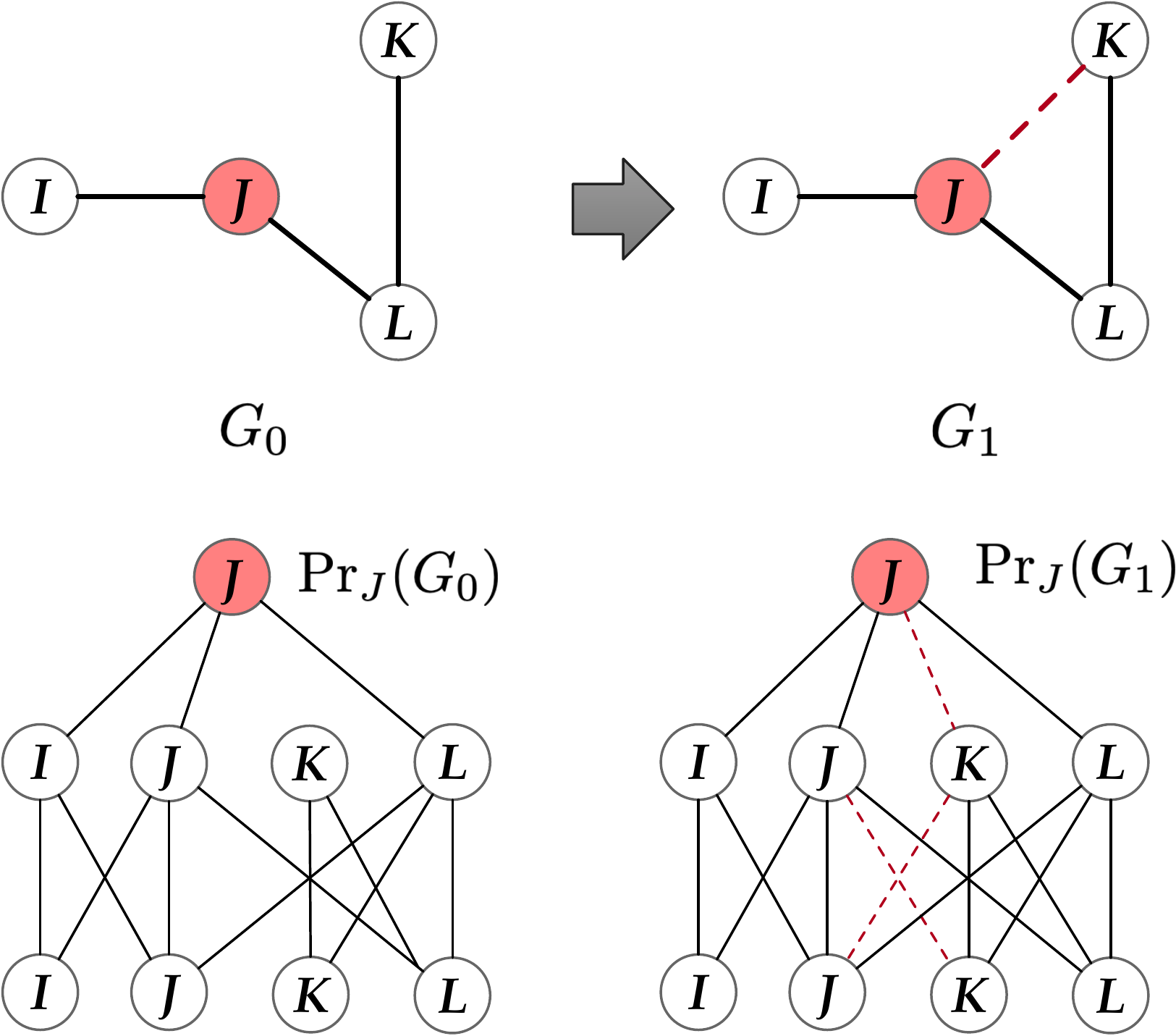}
    \caption{\small
    \textbf{Top}:
    The edge $(J,K)$ is added to the graph $G_0$ to obtain graph $G_1$ and the predicted posterior class distribution of node $J$ changed from $\pr_J(G_0)$ to $\pr_J(G_1)$.
    Prior counterfactual explanation methods can only identify the added edge $(J, K)$ as the cause of the change.
    \textbf{Bottom}:
    the computational graph of the GNN propagates information from leaves to the root $J$. 
    Any paths from any leaf node to the root containing a dashed edge are responsible for the prediction change.
    We further select the few paths that best approximate the change using convex optimization.
    }
    \label{fig:example}
\end{figure}
In the real world, graphs are usually evolving.
For example, pose estimation from videos needs to handle series of graphs representing landmarks of human bodies;
knowledge graphs and social networks are undergoing constant updates.
GNN predictions can change with respect to the evolving graph.
To explain prediction changes is to identify salient graph changes contributing to the prediction changes,
so that humans can understand what causes the change (see Figure~\ref{fig:example}) and how GNN changes its predictions. 

Given a graph $G$ and assuming multi-class prediction, let $\bz_J(G)=[z_1(G),\dots, z_c(G)]$ be the logits of $c$ classes of the node $J$.
The classes are indexed by $j=1,\dots, c$ (in general, we let lower-case $j$ index the neurons of a node).
The predicted class distribution is $\pr_J(G)=\textnormal{softmax}(\bz_J(G))=[\pr_1(G),\dots, \pr_c(G)]$
($\pr_j(G)$ is the probability of class $j$).
On a static graph $G$, $\bz_J(G)$ or $\pr_J(G)$ will be explained~\cite{gnnexplainer,graphlime},
while when $G_0$ evolves to $G_1$,
the difference between $\bz(G_0)$ and $\bz(G_1)$ (or $\pr(G_0)$ and $\pr(G_1)$)
needs to be explained.

Perturbation and counterfactual explanation methods~\cite{Dhurandhar2018,Lucic2021CFGNNExplainerCE}
can attribute a static prediction to
individual edges or nodes by \textit{searching} the optimal change to the input graph, while we are handling \textit{arbitrary} given graph evolution from $G_0$ to $G_1$ and discovering the cause of the change. 
We focus on attributing prediction changes to GNN propagation paths where information flows (see Figure~\ref{fig:example}, top).
The axiomatic attribution method GNN-LRP~\cite{gnnlrp}
handles each class independently given a static graph and cannot explain arbitrary change in the distribution of multiple classes. 
DeepLIFT~\cite{shrikumar17} can explain the log-odd between two classes only (the original and the new predicted classes),
which does not fully capture the change in a multi-class distribution.
For example,
let classes 1 and 2 be the classes with the highest probabilities for the same node $J$ appearing in the graphs $G_0$ and $G_1$, respectively.
One can construct an example multi-class distribution so that $\pr_1(G_0)=0.4$ and $\pr_2(G_1)=0.3$, where the log-odd is positive but the positive contributions of the salient factors to the log-odd cannot explain the reduction in the predicted class probability on $G_1$.

Lastly, to obtain simple explanations, prior methods select a few edges or nodes with the highest importance~\cite{gnnexplainer,shrikumar17,Lundberg2017,gnnlrp}.
However, ranking based on importance can be sub-optimal.
For example, let the change in a logit be 2, which is attributed to three edges as $2=-9+10+1$.
If no more than two edges can be selected,
the simplest and most accurate approximation is $1$,
rather than $(10, -9)$, $(10)$, or $(10, 1)$ found based on the magnitudes of the contributions.

To address the above challenges,
we design an algorithm to explain the change in the probability $\pr(G)$ using a small number of propagation paths that reflect the change of information flow of GNN computation.
We focus on when a single edge is added, with multiple added edges as a simple extension\footnote{Edge removals can be handled similarly, but a mixture of addition and removal need a careful analysis and is left as future work.}.
We first identify the paths containing the added edge,
as other paths make no contributions to the change of prediction.
We attribute the KL-divergence between two multi-class distributions, rather than static or log-odds of posteriors,
to explain the prediction change.
The softmax and the KL-divergence are non-linear everywhere and are thus not amenable to the prior axiomatic attribution methods, which are inherently linear.
Rather, we approximate the KL-divergence in two steps: i) decomposing the changes in the logits over all classes to the altered paths to allow a unique and linear attribution; 
ii) we design a convex optimization problem to select important paths that can optimally approximate the non-linear KL-divergence.

The proposed algorithm goes beyond seeking \textit{what} causes the change in prior works~\cite{gnnexplainer,Lucic2021CFGNNExplainerCE} and explains \textit{how} changing connectivity leads to the change in the information flow of GNN computation graphs.
Theoretically,
we prove that GNN-LRP~\cite{gnnlrp} is a special case of the proposed attribution method starting from a graph without any edge.
Empirically, as prior evaluation metrics do not align well with the quality of the explanations of prediction changes,
we propose a novel metric based on KL-divergence.
We show that solving the convex optimization can select a good portfolio of contributing paths that jointly approximate the change better than ranking-based and linear programming approaches.
\section{Preliminaries and problem definition}

\noindent\textbf{Graph neural networks}.
Assume that we have a trained GNN of $T$ layers that predicts the class distribution of each node $J$ on a graph $G=({\cal V}, {\cal E})$.
Let $\mathcal{N}(J)$ be the neighbors of node $J \in {\cal V}$.
On layer $t$, $t=1,\dots, T$ and for node $J$,
GNN computes $\mathbf{h}_J^{(t)}$ using messages sent from its neighbors:
\begin{eqnarray}
&\bz^{(t)}_J=f_\textnormal{UPDATE}^{(t)}(f_\textnormal{AGG}^{(t)}({\bh^{(t-1)}_J,\bh^{(t-1)}_K,K\in {\cal N}(J)}))  \\
\label{eq:message}
&{\bh}^{(t)}_J=\textnormal{NOLINEAR}({{\bz}}^{(t)}_J) 
\label{eq:GNN}
\end{eqnarray}
$f_\textnormal{AGG}^{(t)}$ aggregates the messages from all neighbors and can be the element-wise sum, average, or maximum of the incoming messages. 
$f_\textnormal{UPDATE}^{(t)}$ maps
$f_\textnormal{AGG}^{(t)}$ to $\bz_J^{(t)}$,
using $\bz_J^{(t)}= \left< f_\textnormal{AGG}^{(t)},\boldsymbol{\theta}^{(t)}\right>$ or a multi-layered perceptron with parameters  $\boldsymbol{\theta}^{(t)}$.
For layer $t\in\{1,\dots,T-1\}$, we let NONLINEAR be the ReLU.
At the input layer, node feature vector $\mathbf{x}_J$ for $J$ is regarded as $\mathbf{h}^{(0)}_J$.
At layer $T$, the logits are $\mathbf{z}_J^{(T)}(G) \triangleq \mathbf{z}_J(G)$, 
whose $j$-th element $z_j(G)$ denotes the logit of the class $j=1,\dots, c$.
$\mathbf{z}_J(G)$ is mapped to the class distribution $\pr_J(G)$ through softmax.
The class $\argmax_j z_j=\argmax_j \pr_j$ is predicted for node $J$.

\noindent\textbf{Evolving graphs}.
Let $G_0=({\cal V}_0,{\cal E}_0)$ denote the initial graph with edges ${\cal E}_0$ and nodes ${\cal V}_0$. Let $G_1=({\cal V}_1, {\cal E}_1)$ denote the graph that evolves from $G_0$.
We assume that 
$G_0$ evolves into $G_1$ with edges added while the nodes ${\cal V}_1={\cal V}_0$ and the node features remain the same.
Let $\Delta E$ denote the set of edges added,
$\Delta {\cal E}=\{e: e \in {\cal E}_1\wedge e \notin {\cal E}_0)\}$.
When the graph $G_0$ evolves to $G_1$, let ${\cal V}^{*}$ denote the set of target nodes, each with a change in its class distribution to be explained:
${\cal V}^{*}=\{J |\pr_j(G_0) \neq \pr_j(G_1), J \in {\cal V}\}$.

\noindent\textbf{Path-based explanations of GNN.}
The GNN prediction at node $J$ is generated by a computation graph,
which is a spanning tree of $G$ rooted at $J$ of depth $T$.
The leaf nodes contain neurons from the input layer ($t=0$) and the root node contains neurons of the output layer ($t=T$) (see Figure~\ref{fig:example}).
The trees completely represent Eq.~(\ref{eq:message}) where messages are passed along the paths traversing from leaves to roots.
Let a path be $(\dots,U,V,\dots,J)$, where
$U$ and $V$ represent any two adjacent nodes on the path and $J$ is designated as the root.
For a GNN with $T$ layers, the paths are sequences of $T+1$ nodes and we let $W(G)$
be the set of all such paths.
Let $W_J(G)\subset W(G)$ be the paths ending at $J$.

The set $\Delta W_J(G_0,G_1)=(W_J(G_1)\setminus W_J(G_0))$
contains all paths with at least one edge that is an added edge,
and therefore 
$\Delta W_J(G_0,G_1)$
completely explains the change in $\pr_J$.
In the example in Figure~\ref{fig:example},
$\Delta W_J(G_0,G_1)=\{(J,K,J),(K,J,J), (K,K,J), (L,K,J),\}$.
However, the contribution of the paths in  $\Delta W_J(G_0,G_1)$ is unknown.
Furthermore, if the paths in $\Delta W_J(G_0,G_1)$ are too many, $\Delta W_J(G_0,G_1)$ is not interpretable.
We define a path-based explanation of a change in $\pr_J$ as a small subset of $\Delta W_J(G_0,G_1)$ closely approximating the change.
Attribution to paths can reveal the details of the change in the information flow of GNN,
which cannot be found by the coarse-grained attribution to altered edges~\cite{Dhurandhar2018,Lucic2021CFGNNExplainerCE}.
\section{AxiomPath-Convex: optimal convex attribution of prediction change to paths}
Most GNN explanation methods attribute a static prediction to nodes or edges on a static graph~\cite{gnnexplainer,gnnlrp}.
Formally,
$\pr_J(G)$ or $\pr_j(G)$ for class $j$
is decomposed into contributions from multiple nodes and edges.
Counterfactual explanations of GNN~\cite{Lucic2021CFGNNExplainerCE} only identify altered edges, rather than changes in the information flow among the neurons, as the cause of the change,
In~\cite{shrikumar17},
the log-odd of the original and new predicted classes $j$ and $j^\prime$ is used to measure the changes in $\pr_J$ when a \textit{fixed} baseline graph $G_0$ evolves to a graph $G_1$.
Besides being unable to handle evolution from an arbitrary $G_0$ to $G_1$,
a more fundamental issue is that the two probabilities $\pr_j(G_0)$ and $\pr_{j^\prime}(G_1)$ are not comparable due to the normalization in softmax and the log-odd does not capture distribution change.
For example,
with three classes, let $\pr_J(G_0)=[0.5, 0.2, 0.3]$ and $\pr_J(G_1)=[0.3, 0.2, 0.5]$.
Then the maximal probabilities are the same for two different classes and the log-odd is $\log \frac{0.5}{0.5}=0$, indicating no change.
In fact, $\pr_j(G_0)$ can be greater than $\pr_{j^\prime}(G_1)$ and the log-odd is positive, giving a wrong sense that class $j$ is more likely than $j^\prime$, although the predicted class changes from $j$ to $j^\prime$.
A metric and an explanation faithfully reflecting the direction and amount of change is yet to be invented.

\subsection{Decomposing KL-divergence to path contributions}
KL-divergence is a well-defined distance metric of probability distributions.
In~\cite{Suermondt1992,gnnexplainer}, KL-divergence is used to measure the approximation quality of a static predicted distribution $\pr_J(G)$.
However, there is no prior work that attributes KL-divergence to paths of information flow of GNN.
We will explain the KL-divergence using paths.

For a target node $J$,
let the difference between its logits
calculated on graphs $G_0$ and $G_1$ be $\Delta \bz_J(G_0, G_1)=\bz_J(G_1)-\bz_J(G_0)=[\Delta z_1,\dots, \Delta z_c]$.
The KL-divergence between two predicted class distributions is
\begin{align}
&\kl{\pr_J(G_1)}{\pr_J(G_0)}\nonumber\\
=&\sum_{j=1}^c \pr_j(G_1) \log[ \pr_j(G_1)/\pr_j(G_0)]\label{eq:sum_of_log_odds}\\
=&\sum_{j=1}^c \pr_j(G_1) [z_j(G_1) - z_j(G_0)]-\log [Z(G_1)/ Z(G_0)],\nonumber\\
=&\sum_{j=1}^c \pr_j(G_1) \Delta z_j-\log [Z(G_1)/ Z(G_0)],\nonumber
\end{align}
where $Z(G_k)=\sum_{j=1}^c\exp(z_j(G_k))$ for $k=0,1$.
In Eq. (\ref{eq:sum_of_log_odds}),
we are adding up the log-odds of the probabilities of the \textit{same} class between two predicted distributions, rather than comparing the probabilities of two \textit{different} classes from two distributions as done by DeepLIFT.

Assume that the change in the logits can be linearly and exactly attributed to $m$ altered propagation paths in $\Delta W_J(G_0, G_1)$ when $G_0$ evolves into $G_1$.
Formally,
$\Delta z_j = \sum_{p=1}^m C_{p,j}$ and 
$\bz_J(G_0)=\bz_J(G_1) - [\sum_{p=1}^{m}C_{p,1},\dots,\sum_{p=1}^{m}C_{p,c}]$,
where $C_{p,j}$ is the contribution of the $p$-th altered path to $\Delta z_j$. 
Note that the paths in the summation $\sum_{p=1}^{m}C_{p,j}$ is the same for all $j=1,\dots,c$.
Accordingly, the KL-divergence becomes
\begin{eqnarray}
    &\sum_{j=1}^c \left[\pr_j(G_1) \sum_{p=1}^{m}C_{p,j}\right]-\log Z(G_1) \nonumber\\
    &+\log \sum_{j^\prime=1}^{c}\exp(z_j(G_1)-\sum_{p=1}^{m}C_{p,j})\label{eq:kl_path} 
\end{eqnarray}
In the next section we will show how to find the $m$ altered paths and calculate $\Delta z_j = \sum_{p=1}^m C_{p,j}$.

In Eq. (\ref{eq:kl_path}),
the number of paths ($m$) can be quite large, especially when adding an edge to a node with high degree,
Using Eq. (\ref{eq:kl_path}) to explain the KL-divergence is not interpretable.
Consider selecting a subset $\Delta E_n$ of $n\ll m$ paths from $\Delta W_J(G_0, G_1)$.
The collection of paths $W_J(G_0)\cup \Delta E_n$ can be considered as the computation graph for node $J$ on a partially evolved graph $G_n$ between $G_0$ and $G_1$.
For example, in Figure~\ref{fig:example},
we can take $E_2=\{(K,J,J), (L,K,J)\}\subset \Delta W_J(G_0, G_1)$ with $n=2$.

\begin{figure}[t]
    \centering
    \includegraphics[width=0.37\textwidth]{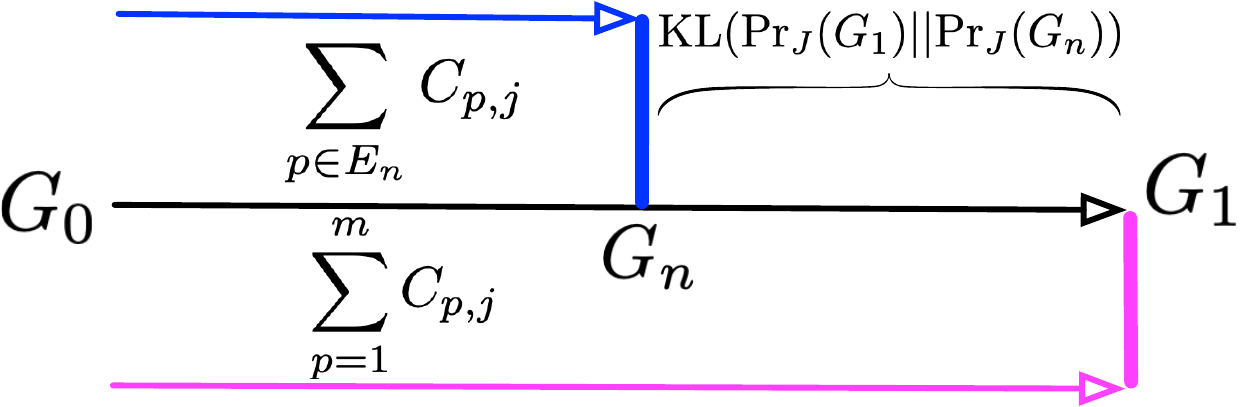}
    \caption{\small Eq. (\ref{eq:kl_path_approx}) equals to 
    $\kl{\pr_J(G_1)}{\pr_J(G_n)}$, which is the residual of approximating the contributions from $m$ altered paths (the purple distance) using $n$ paths from $E_n$ (the blue distance).
    The convex optimization Eq. (\ref{eq:kl_min}) is to minimize the residual.
    }
    \label{fig:kl_path_approx}
\end{figure}
With the partial contributions from the $n$ paths,
the logits on the graph $G_n$ partially evolved from $G_0$ is $\bz_J(G_n)=\bz_J(G_0)+[\sum_{p\in \Delta E_n}C_{p, 0},\dots,\sum_{p\in \Delta E_n}C_{p, c}]$.
Here we abuse the notation $p$ to represent and index a path.
By replacing $G_0$ with $G_n$,
Eq. (\ref{eq:kl_path}) becomes
\begin{eqnarray}
    \sum_{j=1}^c \left[\pr_j(G_1) (z_j(G_1)-z_j(G_0)-\sum_{p\in E_n}C_{p,j}) \right] \nonumber\\
    -\log Z(G_1) +\log \sum_{j^\prime=1}^c \exp(z_{j^\prime}(G_0)+\sum_{p\in E_n}C_{p,j^\prime})\label{eq:kl_path_approx}
\end{eqnarray}
This is the KL-divergence between two distributions $\pr_J(G_1)$ and $\pr_J(G_n)$ and has the minimum of 0.
If it is close to 0, then the contribution from $E_n$ can move $G_0$ to $G_1$ (or $\pr_J(G_0)$ to $\pr_J(G_1)$)
using $n\ll m$ paths, when $m$ paths have significant redundancy.
The idea is shown in Figure~\ref{fig:kl_path_approx}.
It is important to add the contributions to $z_j(G_0)$, $j=1,\dots, c$, to make sure that the added paths will approximate $G_1$ starting from $G_0$.
If the contributions were to be added to a different starting graph (a commonly found example is the empty graph), the KL-divergence in Eq. (\ref{eq:kl_path_approx}) will not explain the change from $G_0$ to $G_1$.

We formulate an optimization problem to select $E_n$. 
Let $x_p\in[0,1]$, $p=1,\dots, m$,
be optimization variables representing the probabilities
of selecting path $p$ into $E_n$.
We solve the following nonlinear minimizing problem:
\begin{eqnarray}
    \bx^\ast&=&\argmin_{\bx\in[0,1]^m} \sum_{j=1}^c \left(-\pr_j(G_1) \sum_{p=1}^{m} x_p C_{p,j}\right)\nonumber\\
    \hspace{.1in}&+&\log \sum_{j^\prime=1}^{c}\exp\left(z_{j^\prime}(G_0)+\sum_{p=1}^m x_p C_{p,j^\prime}\right),
    \label{eq:kl_min}\\
    \textnormal{s.t.} && \sum_{p=1}^m x_p = n. 
\end{eqnarray}
The first term is linear in $\bx$ and the second term is the composition of a log-sum-exp function (convex and non-decreasing) and a linear function of $\bx$. The constraint function is linear in $\bx$. Therefore, the optimization problem is convex and has a unique optimal solution given the constant $n$.
In going from Eq. (\ref{eq:kl_path_approx}) to the objective, we ignore the constants $z_j(G_0)$, $z_j(G_1)$, and $Z(G_1)$.
The linear constraint is to ensure the total probabilities of the selected edges is $n$.
Note the similarity between the optimization problem and the minimization of multi-class logistic regression: $x_p$ can be regarded as regression parameters and the $m$ paths as features.
After obtaining $\bx^\ast$, we rank the paths based on $\bx^\ast$ and include the top $n$ paths in $E_n$.

\subsection{Background on DeepLIFT}
It remains to calculate $C_{p,j}$ for all paths $p$.
We extend DeepLIFT to attribute prediction change to $C_{p,j}$ of the paths in $\Delta W_J(G_0, G_1)$.
Let
the neuron $h^{(t+1)}$ be computed by
$h^{(t+1)} = f(h^{(t)}_1,\dots,h^{(t)}_n)$,
with inputs neurons $h^{(t)}_1,\dots,h^{(t)}_n$. Given the \textbf{reference activations} $h^{(t)}_1(0),\dots,h^{(t)}_n(0)$ at layer $t$, we can calculate the reference activation $h^{(t+1)}(0)=f(h^{(t)}_1(0),\dots,h^{(t)}_n(0))$.
The \textbf{difference-from-reference} is $\Delta h^{(t+1)} = h^{(t+1)}(1)-h^{(t+1)}(0)$ and $\Delta h^{(t)}_i=h_i^{(t)}(1)-h^{(t)}_i(0)$, $i=1,\dots, n$.
The 0 in parentheses indicates the original input and the 1 indicates the changed input.
The contribution of neuron $h^{(t)}_i$ to $\Delta h^{(t+1)}$ is denoted by $C_{\Delta h^{(t)}_i \Delta h^{(t+1)}}$ such that $\sum_{i=1}^n C_{\Delta h^{(t)}_i \Delta h^{(t+1)}}=\Delta h^{(t+1)}$ and $\Delta h^{(t+1)}$ is preserved by the sum of all contributions. 



\subsection{DeepLIFT for GNN}
\begin{figure}[t]
    \centering
    \includegraphics[width=0.45\textwidth]{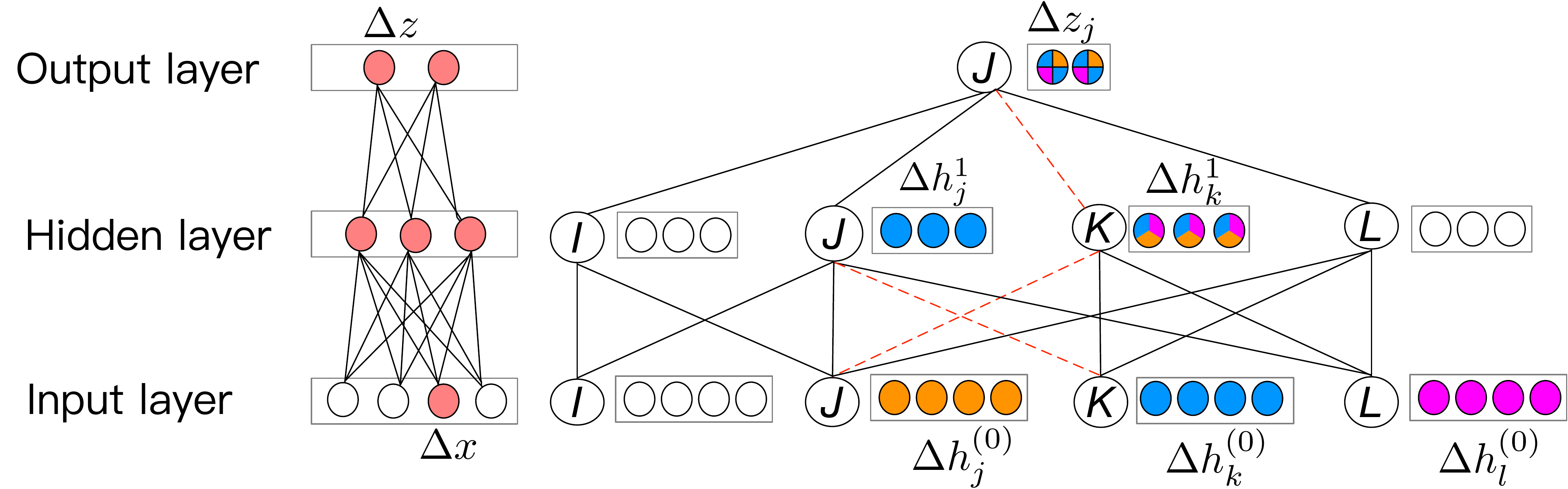}
    \caption{\small
    Circles in rectangles are neurons, and a neuron is colored if there is a non-zero contribution to the prediction change.
    \textbf{Left}:
    DeepLIFT finds the contribution of an input feature to the change in an output neuron in deep neural networks. It cannot deal with altered paths in GNN.
    \textbf{Right}:
    A two-layer GNN for node classification: the four colored quadrants in $\Delta z_j$ at the output layer can be traced back to the non-zero changes in the contribution of the input neurons to $J$ at the root.
    For example, the two blue quadrants at $J$ at the output layer can be traced back to the blue neurons in node $K$ at the input layer through paths $(K, K, J)$ and $(K, J, J)$.
    The details of the tracing can be found in the main texts.
    }
    \label{fig:difference}
\end{figure}

We let the reference activations be computed by the GNN on the graph $G_0$.
We assume that there is just one edge  added to $G_0$ to obtain $G_1$ since the more general case with multiple added edges can be decomposed into multiple steps of adding one edge. Let $U$ and $V$ represent any adjacent nodes in a path, where $V$ is closer to the root $J$.


\noindent\textbf{Difference-from-reference of neuron activation and logits}. 
For a path $p$ in $\Delta W_{J}(G_0,G_1)$, let $p[t]$ denote the neurons of the node at layer $t$. For example, if $p=(I,\dots,J)$, $p[T]$ represents the neurons of node $J$ at layer $T$ and $p[0]$ represents the neurons of node $I$ at layer $0$.
Given a path $p$,
let $\bar{t}=\max\{\tau |p[\tau]=V\textnormal{ and } 
p[\tau-1]=U\textnormal{ and } (U,V) \textnormal{ is a newly added edge}\}$.
 When $t \geq \bar{t}$, the reference activation of $p[t]$ is $h_{p[t]}^{(t)}(G_0)$. 
 While when $t< \bar{t}$,
 the reference activation of $p[t]$ is zero, because the message of $p[t]$ cannot be passed along the path $(p[t],\dots,J)$ to $J$ in $G_0$,
 since the edge $(U,V)$ must be added to $G_0$ to connect $p[t]$ to $J$ in the path.
 We thus calculate the difference-from-reference of neurons at each layer as follows: 
\begin{eqnarray}
\Delta h_{p[t]}^{(t)}=\left\{
\begin{array}{lr}
    h_{p[t]}^{(t)}(G_1)-h_{p[t]}^{(t)}(G_0) & t\geq \bar{t}, \\
    h_{p[t]}^{(t)}(G_1)&  \textnormal{otherwise.}
\end{array}
\right.
\label{eq:difference}
\end{eqnarray}

For example, in Figure~\ref{fig:difference}, the added edge is $(J,K)$.
For the path $p=(K,K,J)$ in $G_1$, $U=K,V=J$. 
$\bar{t}=2$ and $\Delta h_k^{(0)}=h_{k}^{(0)}(G_1)$, 
because in $G_0$, the neuron $k$ at layer $0$ cannot pass message to the neuron $j$ at the output layer along the path $(K,K,J)$ in $G_0$.

The change in the logits $\Delta z_{p[t]}^{(t)}$ can be handled similar to $\Delta h_{p[t]}^{(t)}$.

\noindent \textbf{Obtain  multiplier  of  each  neuron to its immediate successor.} 
The linear rule and the rescale rule designed by the DeepLIFT can be used for backpropagation to obtain the multiplier  of  each  neuron to its immediate successor.  

We choose element-wise sum as the $f_\textnormal{AGG}$ function to ensure that the attribution by DeepLIFT can preserve  the  total  change  in  the  logits such that $\Delta z_j= \sum_{k=1}^{m} C_{p,j}$. 
Then, $z_{v}^{(t)}$ is rewritten into the following form. 
\begin{eqnarray}
z_{v}^{(t)}=\sum_{U \in N(V)}\left(\sum_{u \in U} h_{u}^{(t-1)} \theta_{u, v}^{(t)}\right)
\end{eqnarray}
where $\theta_{u,v}^{(t)}$ denotes the element of the matrix $\theta ^{(t)}$ that links neuron $u$ to neuron $v$.
Therefore, we can obtain the multiplier of the neuron $u$ to its immediate successor neuron $v$. 
\begin{eqnarray}
m_{\Delta h^{(t-1)}_u \Delta h^{(t)}_v}=\frac{\Delta h^{(t)}_v}{\Delta z^{(t)}_v } \times \theta_{u,v}^{(t)}
\label{eq:multiplier}
\end{eqnarray}

Note that the output of GNN model is $z_j^{t}$, thus $m_{\Delta h^{(T-1)}_{p[t-1]} \Delta z_j}=\theta^{(T)}_{p[t-1],j}$. We can obtain the multiplier of each neuron to its immediate successor in the path according to Eq.~(\ref{eq:multiplier}) by letting $t$ go from $T$ to $1$.
After obtaining the $m_{\Delta h_{i}^{(0)} \Delta h_{p[1]}^{(1)}},
\dots, m_{\Delta h_{p[T-1]}^{(T-1)}, \Delta z_{j}}$, according to the chain rule, we can obtain $m_{\Delta h_{i}^{(0)} \Delta z_j}$.

\noindent \textbf{Calculate the contribution of each path.} 
For the path $p$ in $\Delta W_{J}(G_0,G_1)$, we obtain the contribution of the path by summing up the input neurons' contributions: $C_{p,j}=\sum \limits_{i} m_{\Delta h_i^{(0)} z_j} \times h_i^{(0)}$,
where $i$ indexes the neurons of the input (a leaf node in the computation graph of the GNN).

\begin{thm}
\label{thm:ranking_Pareto}
The GNN-LRP is a special case if the reference activation is set to the empty graph. 
\end{thm}
See the Appendix for a proof. 

\section{Experiments}
\begin{table}[t]
    \scriptsize
    \caption{\small Datasets used in experiments.
    More details in Appendix.}
    \centering
    \begin{tabular}{c|c|c|c|c|c}
    \toprule
    \textbf{Datasets} & \textbf{Classes} & \textbf{Nodes} & \textbf{Edges} & \textbf{Edge/Node} & \textbf{Features}\\
    \midrule
    \textbf{Cora} & 7 & 2,708 & 10,556 & 3.90 & 1,433\\
    \textbf{Citeseer} & 6 & 3,321 & 9,196 & 2.78 & 3,703\\
    \textbf{PubMed} & 3 & 1,9717 & 44,324 & 2.24 & 500\\
    \midrule
    \textbf{Amazon-C} & 10 & 13,752 & 574,418 & 41.77 & 767\\
    \textbf{Amazon-P} & 8 & 7,650 & 287,326 & 37.56 & 745\\
    \midrule
    \textbf{Coauthor-C} & 15 & 18,333 & 327,576 & 17.87 & 6,805\\
    \textbf{Coauthor-P} & 5 & 34,493 & 991,848 & 28.76 & 8,415\\
    \bottomrule
    \end{tabular}
     \label{tab:datasets}
\end{table}

\begin{figure*}[ht]
    \centering
  \subfloat{\includegraphics[width = 0.25\textwidth]{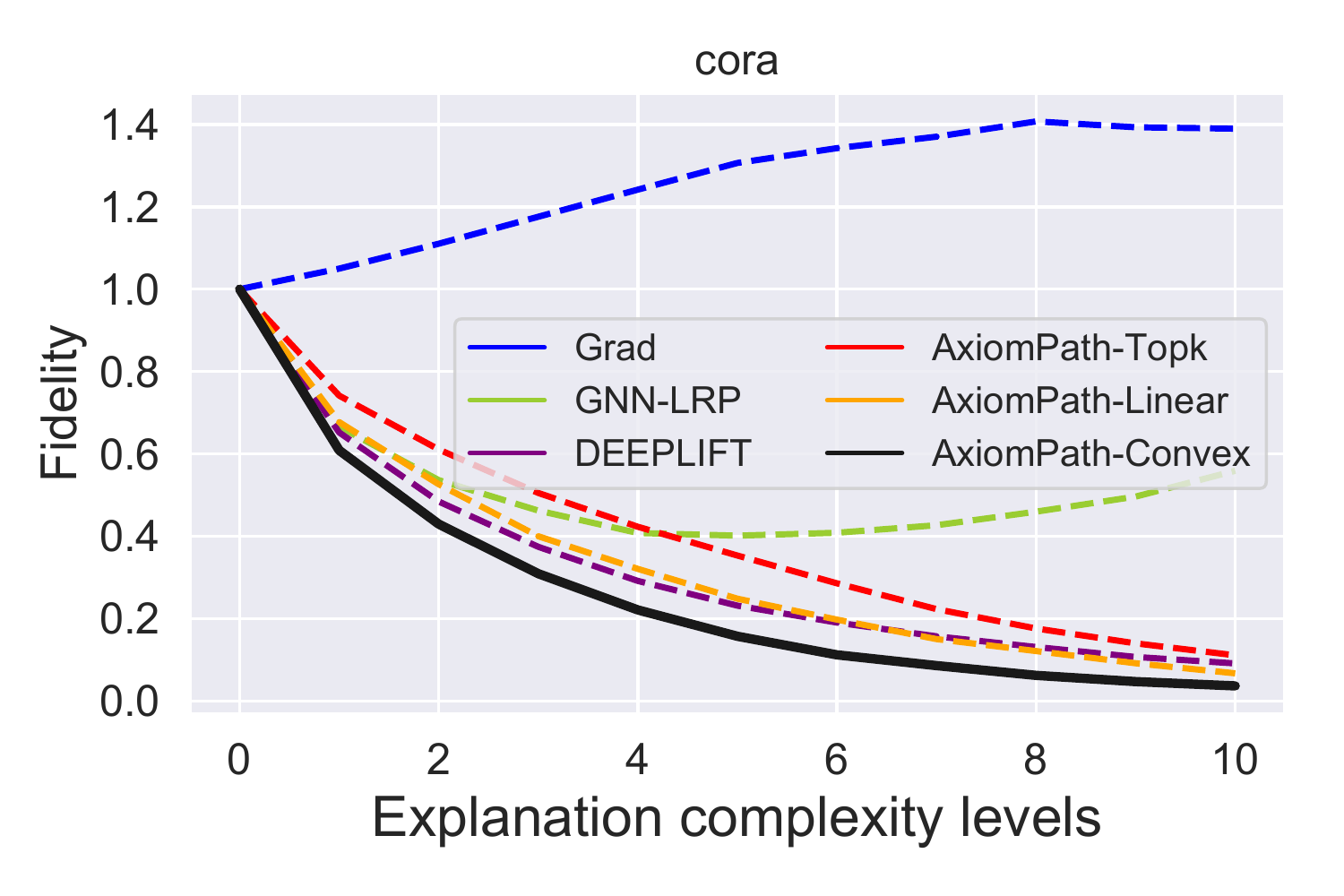}
  }
  \subfloat{\includegraphics[width = 0.25\textwidth]{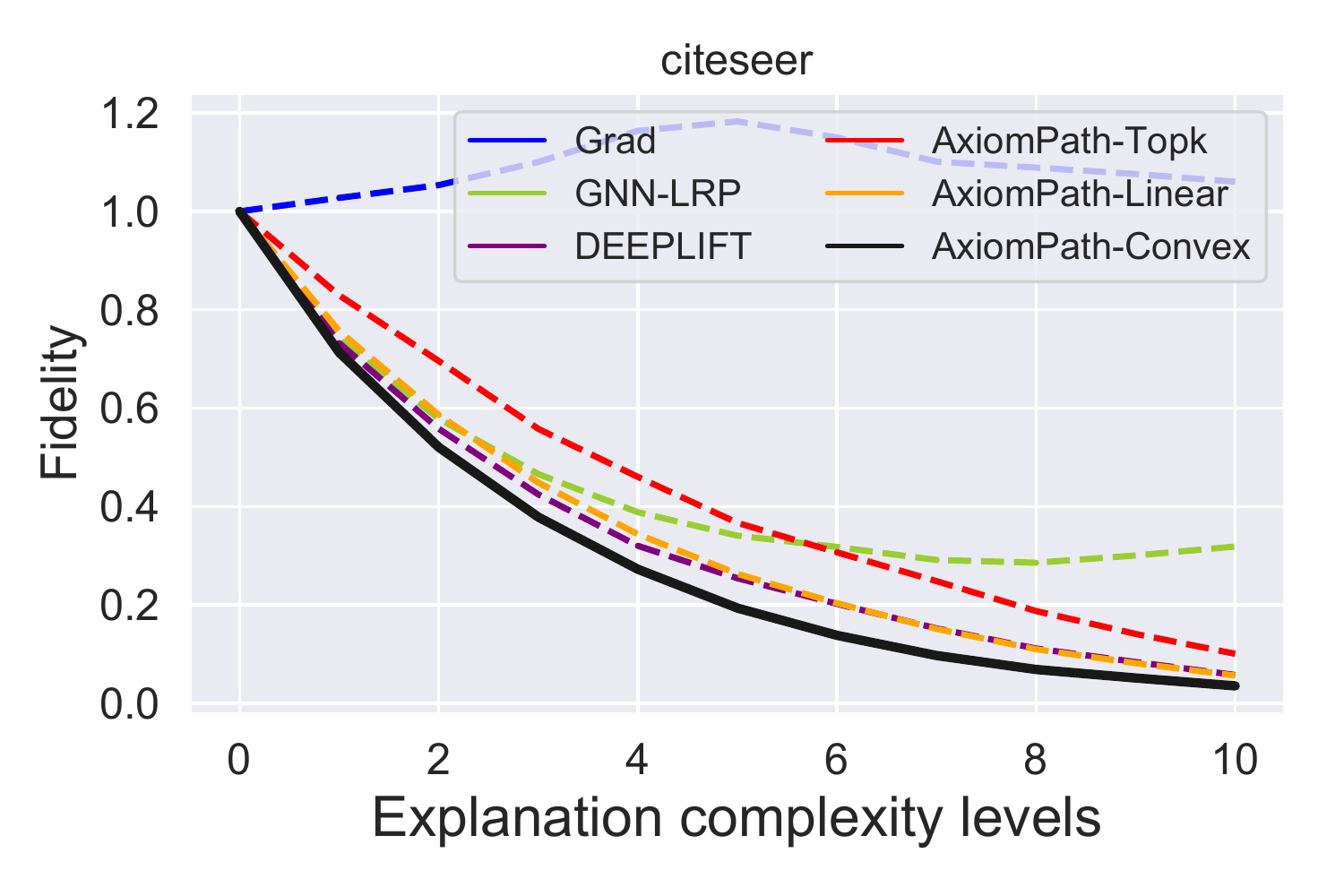}
  }
  \subfloat{\includegraphics[width = 0.25\textwidth]{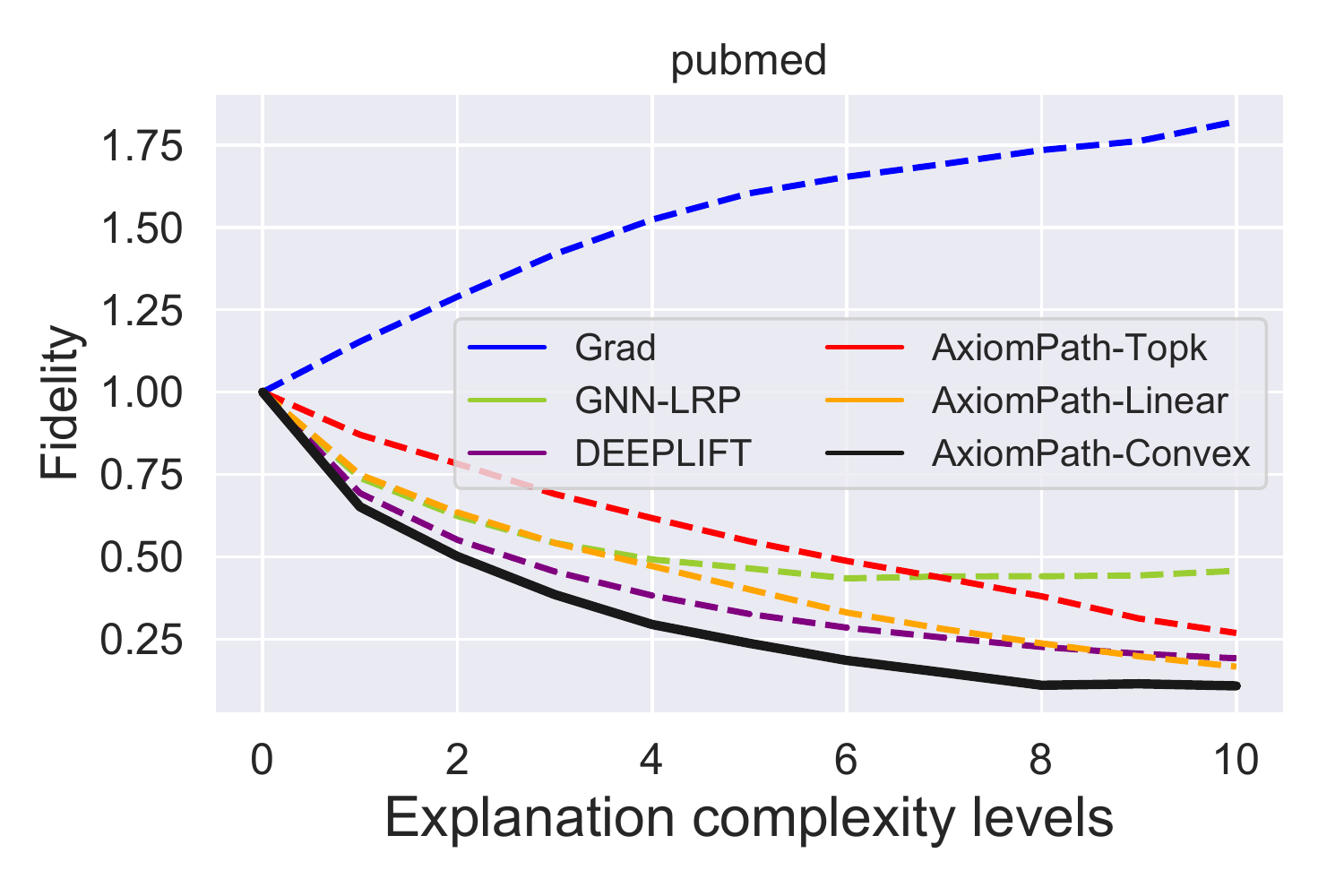}
  }\\
  \subfloat{\includegraphics[width = 0.25\textwidth]{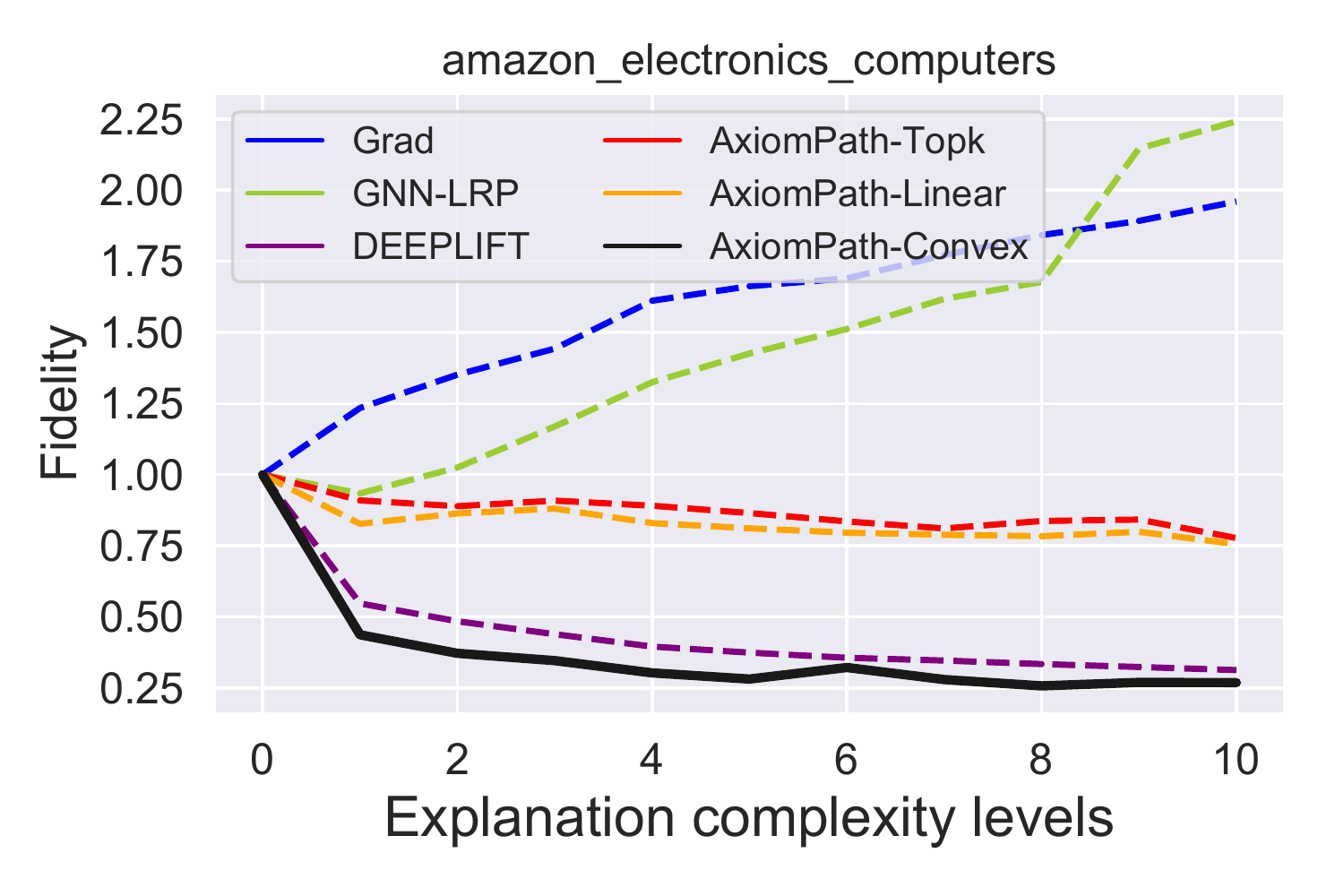}
  }
  \subfloat{\includegraphics[width = 0.25\textwidth]{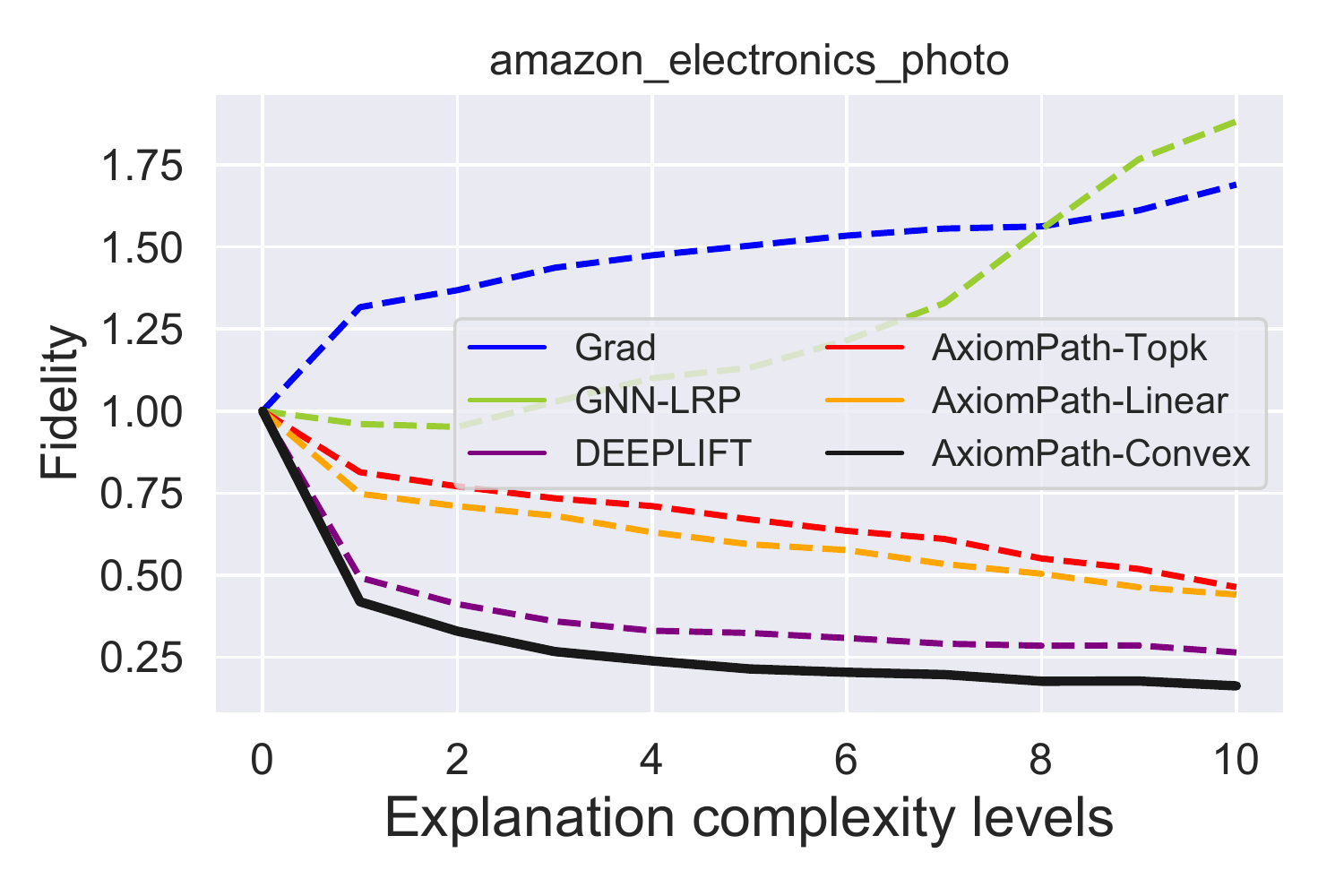}
  }
  \subfloat{\includegraphics[width = 0.25\textwidth]{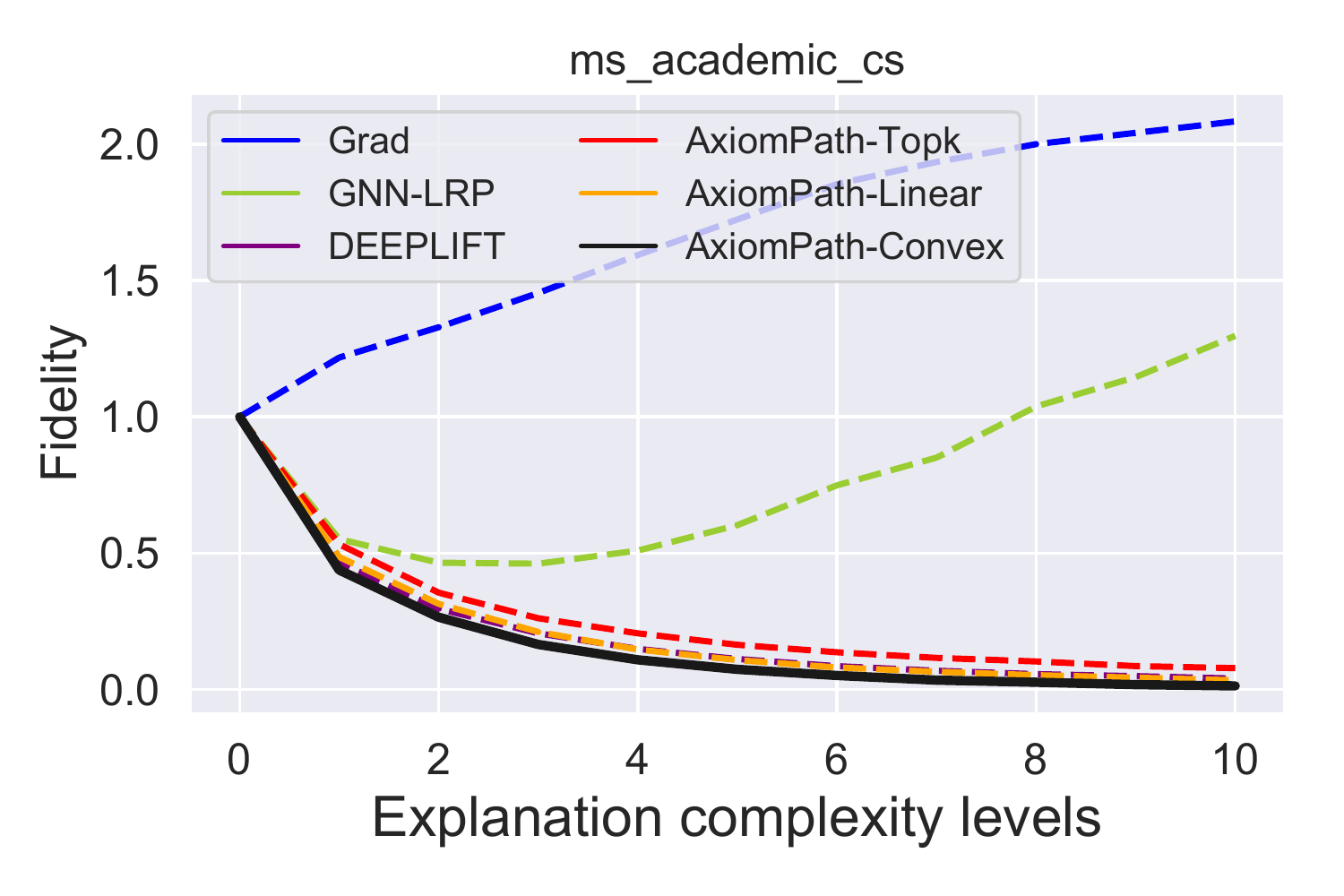}
  }
  \subfloat{\includegraphics[width = 0.25\textwidth]{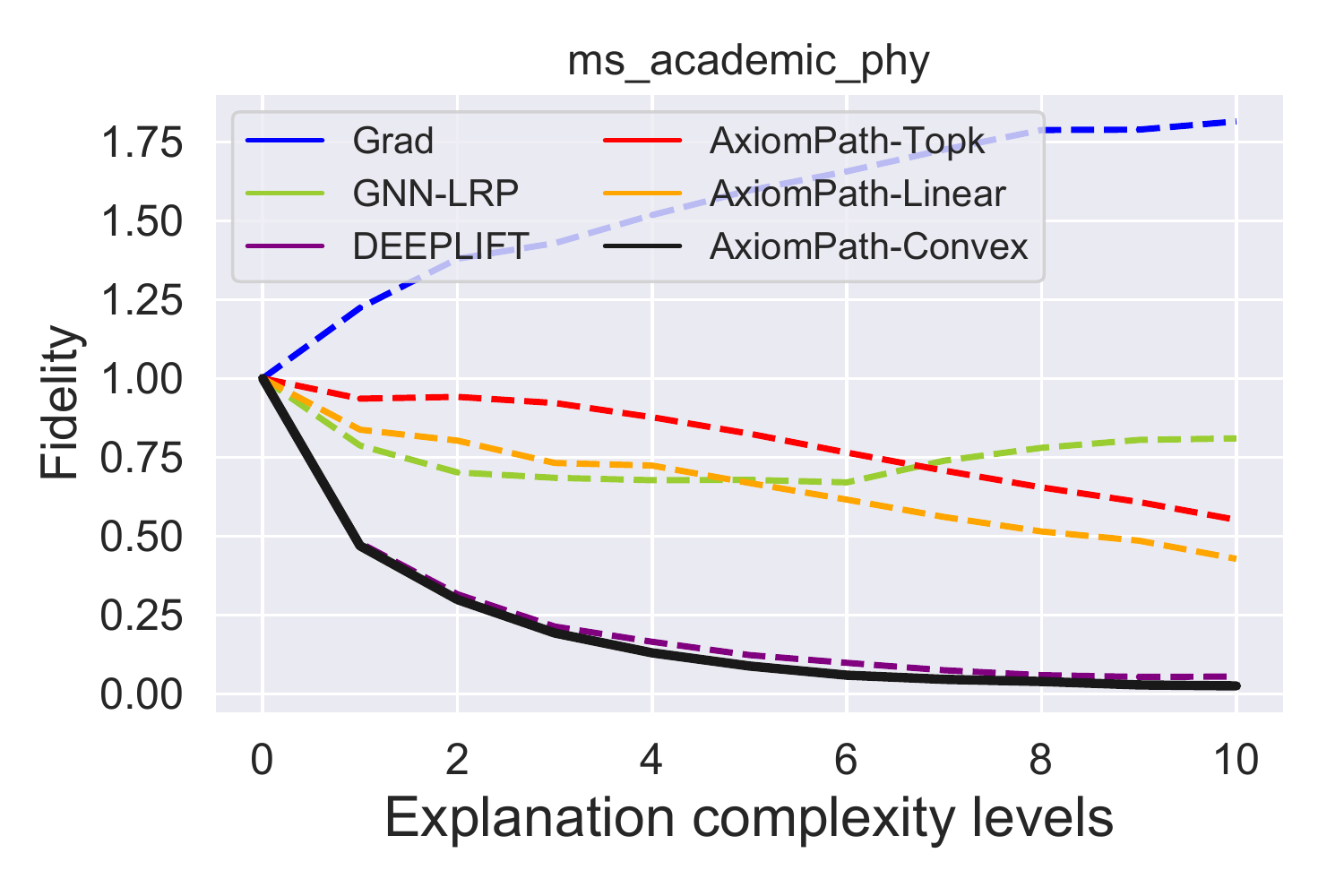}
  }
  \vspace{-.1in}
  \caption{\small Mean \textbf{Fidelity}$_\textnormal{KL}^{-}$ over all datasets.
  Standard deviations are in the Appendix. AxiomPath-Convex has the best performance.
  }
\label{fig:overall}
\end{figure*}
\noindent\textbf{Datasets and tasks}.
We select graph datasets that are suitable for node classification.
Certain datasets used in prior GNN explanation works are not applicable.
For example, it is hard to create evolving molecule graphs~\cite{mutag,tox21} and dependency parse trees~\cite{sst2,gnnlrp} while maintaining the syntactic integrity of the graphs.
Besides, these datasets are majorly used for graph classification rather than node classification.
Second,
the motif-based patterns (\textit{grids} vs. \textit{trees}) of the synthetic graphs~\cite{gnnexplainer} cannot be explained by simple paths.
Therefore, we adopt 7 graph datasets from three real-world applications for evaluation (see Table~\ref{tab:datasets} and the Appendix for the details).

\noindent\textbf{Baselines}.
We adopt the following methods as baselines. 

\begin{itemize}[leftmargin=*]
 \item \noindent\textbf{Gradient} (Grad) employs the gradients of the logits of class $j$ of target node $J$ with respect to individual edges
 as edge importance.
 Path importance is the sum of absolute gradients of the edges of a path.
 Paths are ranked and selected from the existing and newly added paths.
\item \noindent\textbf{GNN-LRP} adopts the back-propagation attribution method LRP to GNN~\cite{gnnlrp}.
However, it only attributes the single predicted class probability $\pr_j(G_1)$ to input neurons and the original graph $G_0$ is not accounted for.
Path relevance is calculated in the same way as specified in~\cite{gnnlrp} and the path ranking and selection is done as in Grad.
\item \textbf{DeepLIFT}~\cite{shrikumar17}
is the most relevant baseline as it can explain the difference between the predicted class probabilities from two graphs.
It can handle zero activations ignored by GNN-LRP. 
However, it attributes the log-odds between the original and new predicted classes to neurons and does not directly apply to GNN. 
We design the GNN version of DeepLIFTGNN by branching out to all neurons of all neighbors of a node during back-propagation.
Then we take the sum of contributions of the input neurons at node $I$ to class $j$
as the contribution of the unique path from $I$ to $J$ for that class. 
Lastly, the difference between each path's contributions to the new and original predicted classes is used to rank and select paths in $\Delta W_J(G_0, G_1)$.
\item \textbf{AxiomPath-Topk} is a variant of AxiomPath-Convex,
without solving the convex programming.
Rather, it selects paths with top contributions to all classes $\sum_{j=1}^{c}C_{p, j}$.
The path ranking and selection are the same as DeepLIFT.
\item \textbf{AxiomPath-Linear} solves the optimization problem in Eq. (\ref{eq:kl_min}) without the log term, leading to a linear programming problem.
The resulting optimal $\bx^\ast$ is processed in the same way as AxiomPath-Convex.
\end{itemize}

Certain prior methods are not suitable as baselines.
CAM and Grad-CAM are not applicable since they cannot explain node classification models~\cite{Yuan2020ExplainabilityIG}.
PGExplainer and SubgraphX find subgraphs as explanations of static predictions while being not comparable with path-based explanations.
GNNExplainer finds the importance of individual edges that multiplex an edge's contributions to multiple paths of information flow. Simply combining edge importance to select paths leads to much worse fidelity and thus GNNExplainer is not a relevant baseline.

\noindent\textbf{Experimental setup}.
For each graph dataset, we first train and fix a GNN for node classification on the labeled training set.
Randomly selected edges are added to simulate the graph evolution. 
At the end of the evolution,
the set of target nodes ${\cal V}^{*}$ with a different prediction is collected and the change in each target node is explained using the various methods.
The target nodes are grouped based on the number of added paths for the results to be comparable.
For each group, we let $n$ (the number of selected edges) range in a pre-defined set to create 10 levels of explanation complexity ($|E_n|$).
We use SCS solver in the cvxpy library to solve the constrained convex optimization problem. 
We run the evolution and explanation for $N$ times to calculate the means and standard deviations of the following fidelity metric over target nodes. (see the Appendix for more details)

\noindent\textbf{Evaluation Metrics}.
Even though explanation visualization allows humans to judge whether the explanations are reasonable,
it cannot objectively evaluate the explanations.
Instead, we design the following new metric to evaluate fidelity in explaining changes in class distributions:
\begin{equation}
\mathbf{Fidelity}_{\textnormal{KL}}^{-}=
\frac{\kl{\pr_J(\neg G_n)}{\pr_J(G_0)}}{\kl{\pr_J(G_1)}{\pr_J(G_0)}}\nonumber
\end{equation}
where $\pr_J(\neg G_n)$ is the class distribution computed on the computation graph for $G_1$ with the selected paths $E_n$ \textit{removed}.
Intuitively,
if $E_n$ indeed contains the important paths to turn $G_0$ into $G_1$,
the less information the remaining paths can propagate, the more similar is $\neg G_n$ to $G_0$, and the smaller the ratio.
At one extreme,
$E_n=\emptyset$ so that the numerator is just $\kl{\pr_J(G_1)}{\pr_J(G_0)}$ (nothing is removed) and the ratio is 1.
At the other extreme, $E_n$ includes all added paths and $\neg G_n$ degrades to $G_0$ and the ratio is 0.
The denominator is for normalization since the total change can be of different scales for different $J$.  
We intentionally design this metric to differ from the objective function in Eq. (\ref{eq:kl_min}) so that AxiomPath-Convex has less privilege.
For fairness, we ensure the same number of edges are removed from the same set of paths, so that the number of remaining paths is the same for all methods.

\subsection{Performance evaluation and comparison}
\noindent\textbf{Path importance on real-world datasets.}
In Figure~\ref{fig:overall}, we demonstrate the effectiveness of the salient path selection of AxiomPath-Convex.
For each dataset, we report the mean fidelity over 10 random experiments and over target nodes whose predictions are altered (standard deviations are reported in the Appendix).
The lower the fidelity, the better (since we intend to remove the important paths identified by $E_n$).
From the figure, we can see that AxiomPath-Convex has the smallest fidelity
over all levels of explanation complexities and over all datasets.
On four datasets (Cora, Citeseer, Amazon-C, and Amazon-P), the gap between AxiomPath-Convex and the runner-up (DeepLIFT) is significant.
On the remaining three datasets, the gap is less significant but still not ignorable.
The third best baseline is AxiomPath-Linear, indicating the power of using optimization for selecting explaining paths, and more importantly, the need to approximate the KL-divergence in full using non-linear convex optimization.
GNN-LRP lacks consistency: on all datasets, it has increasing fidelity when the complexity level goes up after some points.
AxiomPath-Topk underperforms AxiomPath-Linear, indicating that paths must be selected jointly rather than independently.
Furthermore, though descending at a slower pace, AxiomPath-Topk and AxiomPath-Linear consistently decreases the fidelity as more important paths are identified, indicating the ``consistency'' of the proposed attribution method shared by the AxiomPath-$\ast$ family.
Grad always (and sometimes for GNN-LRP) fails to find the right paths to explain the change, as they are designed for static graphs.

\noindent\textbf{Preserving changes during attributions.}
\begin{figure}[t]
    \centering
  \includegraphics[width = 0.32\textwidth]{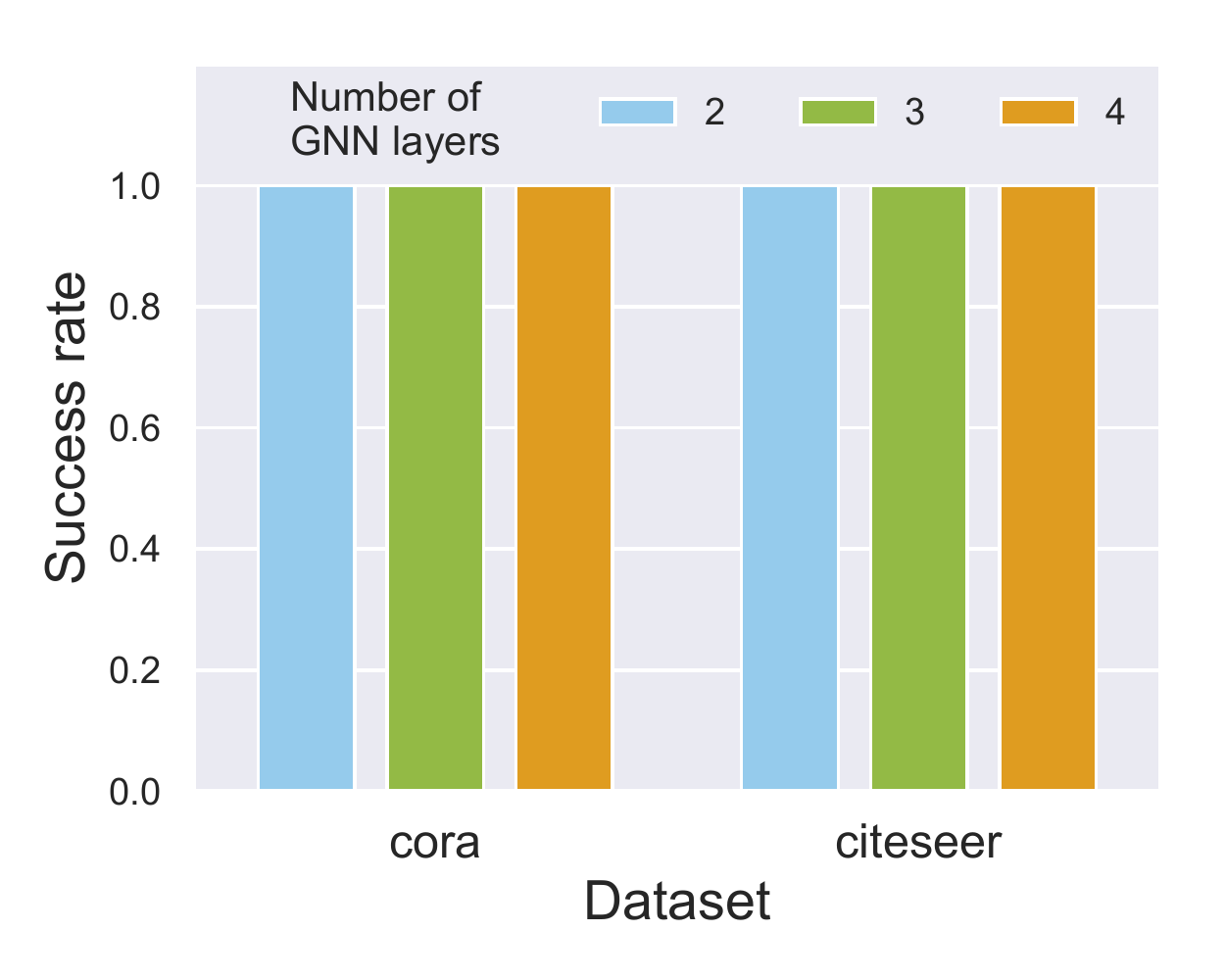}
  \caption{\small Success rate of preservation on Cora and Citeseer datasets for three GNN models with different layers. 
  }
\label{fig:property}
\end{figure}
Edges randomly selected are added or deleted to simulate the graph evolution and the parameters for the GNN model are randomly generated. For the target node in ${\cal V}^*$, if the difference $|\Delta z_j-\sum_{p=1}^{m} C_{p,j}|$ is less than $10^{-5}$, we regard the change as successfully preserved. We calculate the percentage of successful preservation among ${\cal V}^\ast$.
Figure~\ref{fig:property}) shows that the attribution preserves the change in the logits 100\%.

\begin{figure}[htbp]
    \centering
    \subfloat{\includegraphics[width=0.25\textwidth]{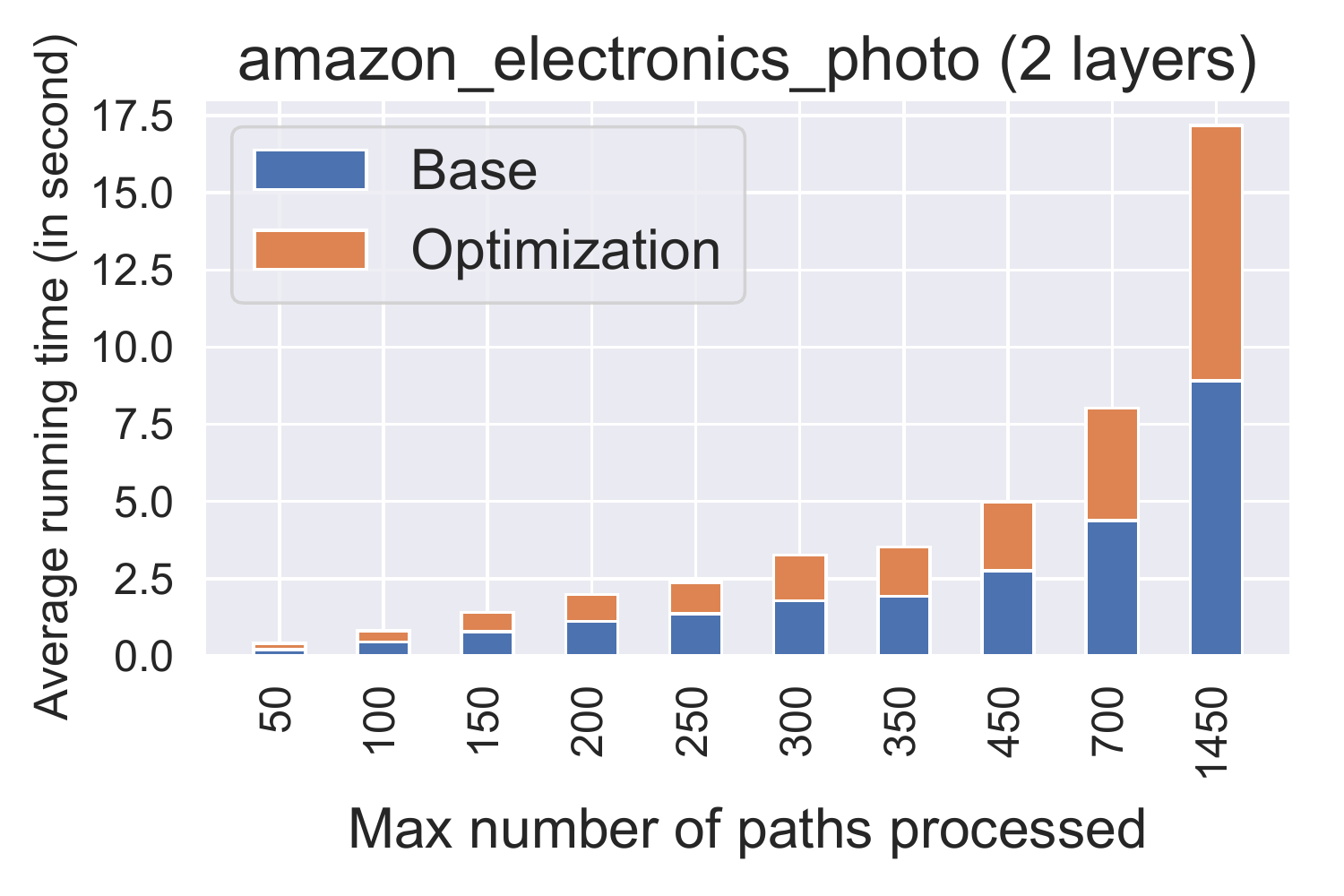}}
    \subfloat{\includegraphics[width=0.25\textwidth]{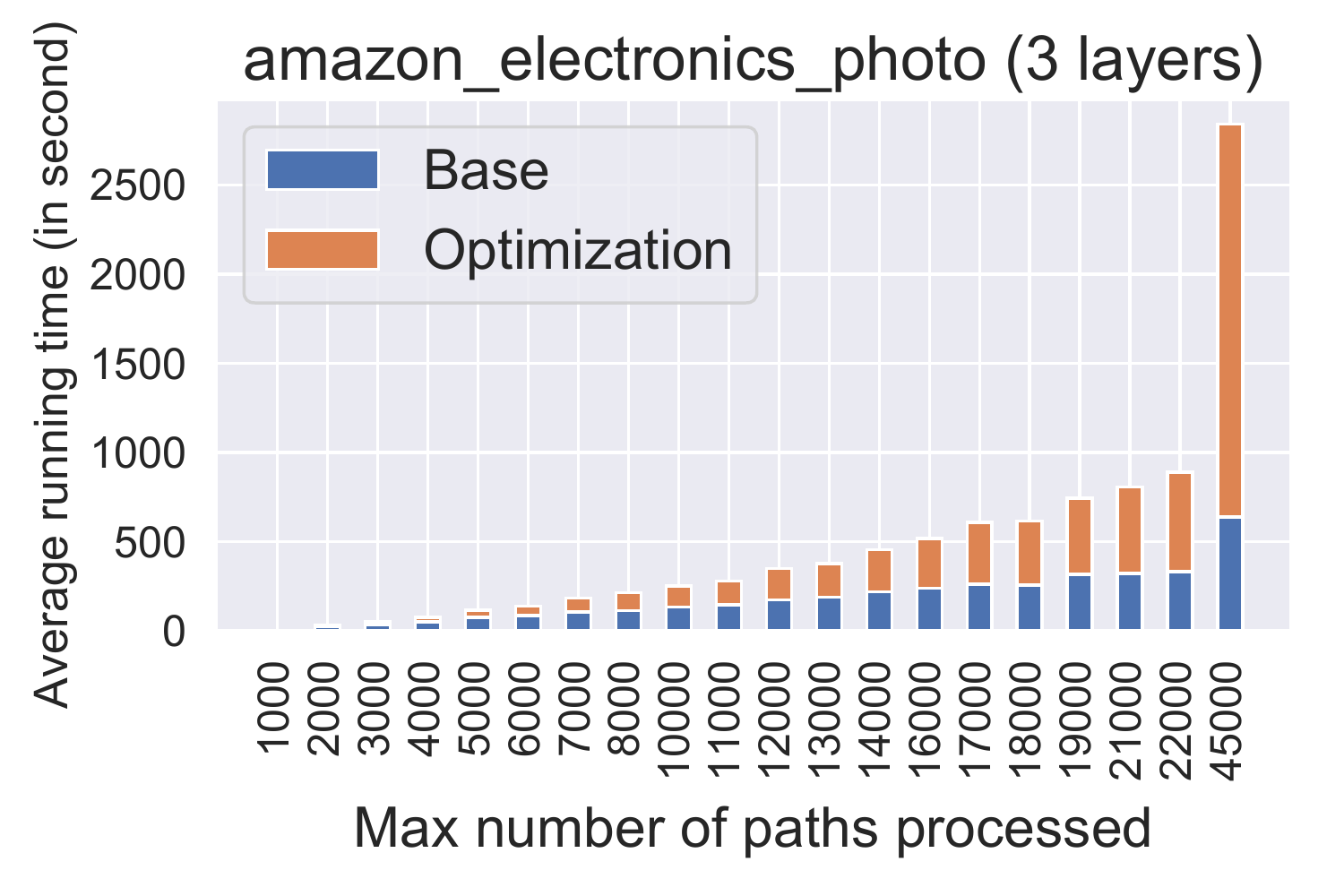}}\\
    \subfloat{\includegraphics[width=0.25\textwidth]{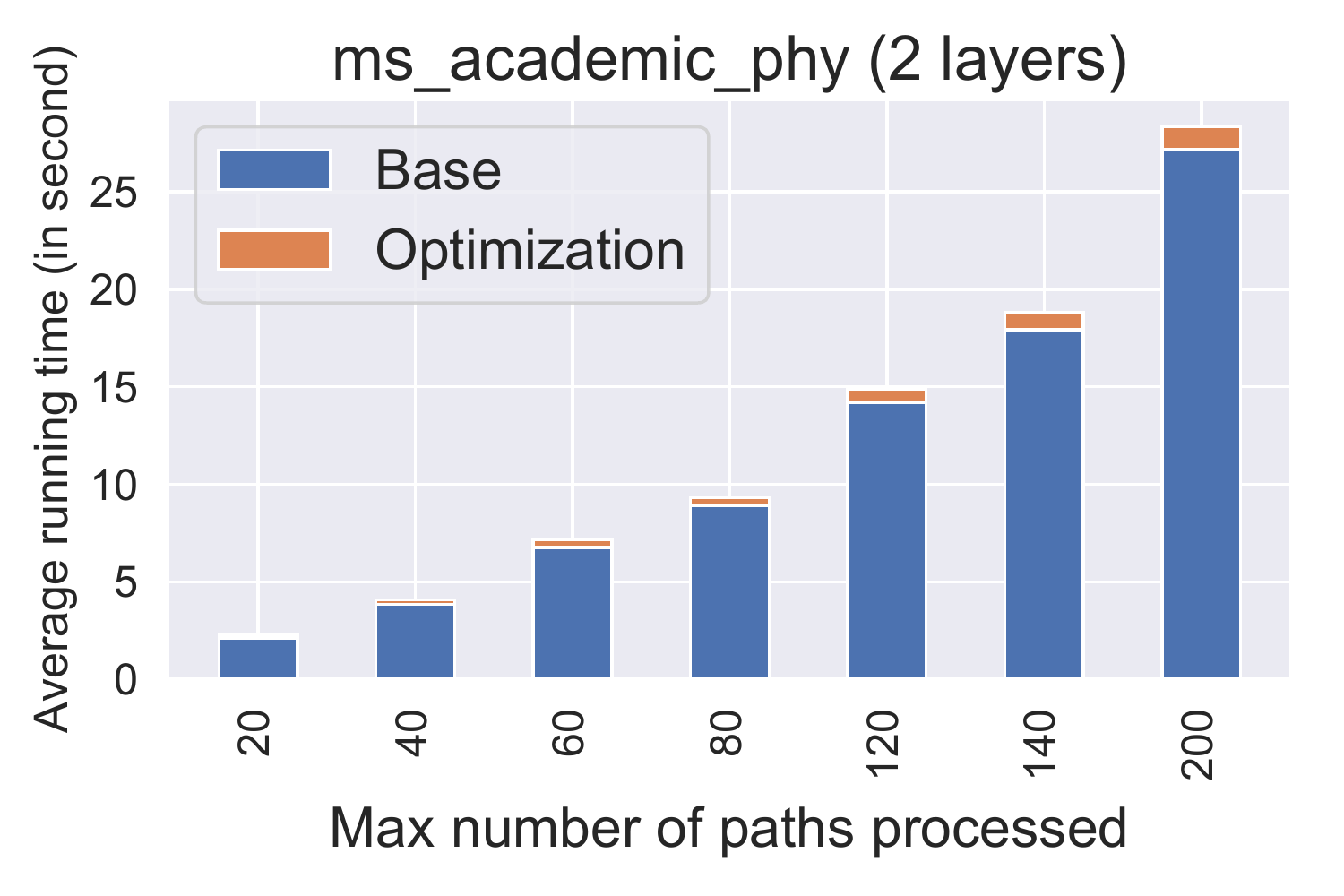}}
    \subfloat{\includegraphics[width=0.25\textwidth]{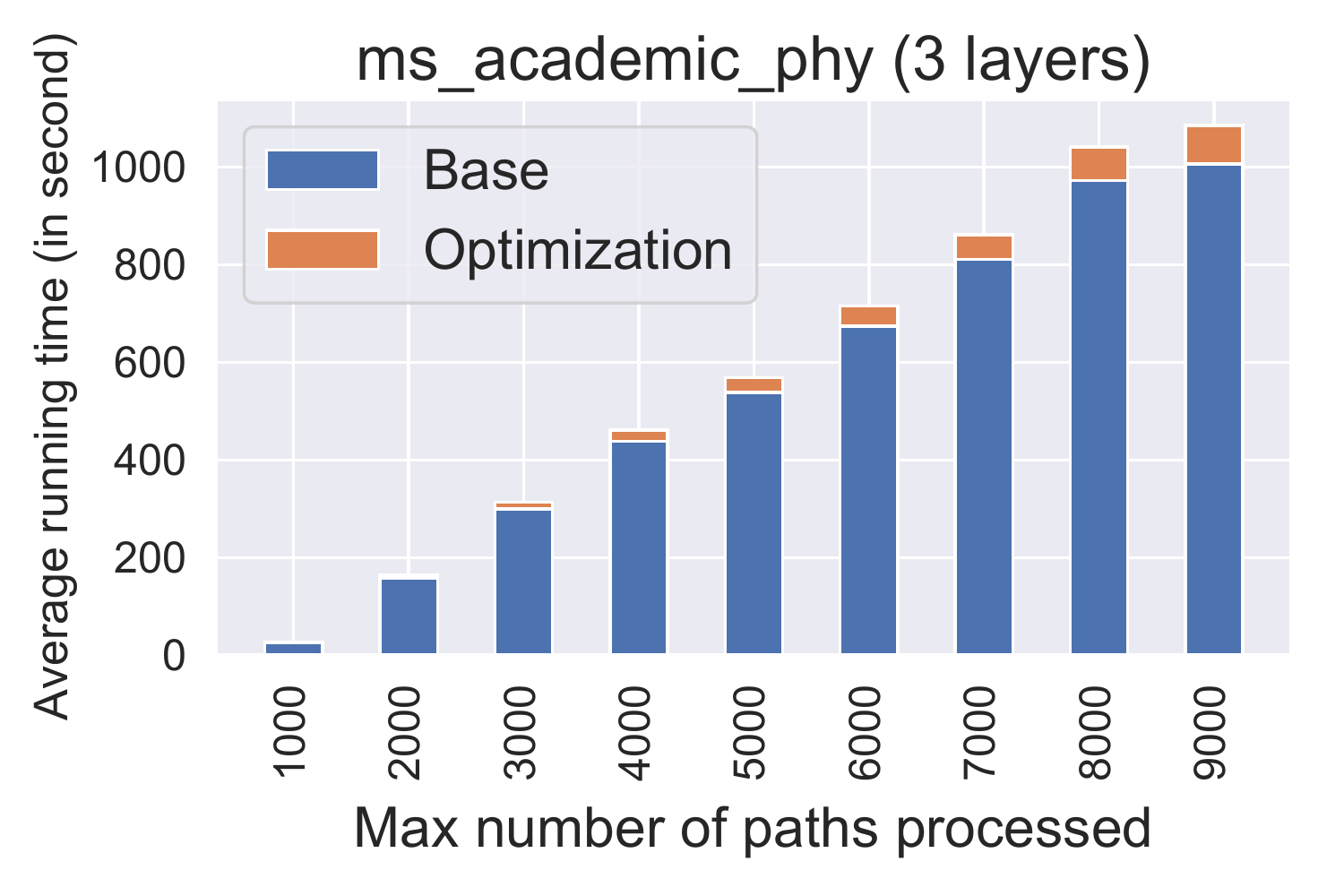}}
    \caption{\small Running time of optimization.}
    \label{fig:opt_running_time}
\end{figure}

\noindent\textbf{Running time overhead of convex optimization.}
We plot the base running time of searching paths in $\Delta W(G_0, G_1)$
and attribution \textit{vs.} the running time of optimization afterwards.
From Figure~\ref{fig:opt_running_time},
we can see that in most cases when the number of new paths is not extremely large, the optimization step takes a reasonable amount of time compared to the base running time.
However, there are two exceptions on the Amazon dataset when the number of new paths is large.
We argue that's because the off-the-shelf optimizer in scipy is used and a customized optimizer can reduce the running time, by early stopping.
\section{Related work}
Regarding what about GNN is explained,
there are methods explaining the predicted class distribution of graphs or nodes using mutual information as the objective function~\cite{gnnexplainer}.
Other works explained the logit or probability of a \textit{single} class~\cite{gnnlrp}.
CAM, GradCAM, GradInput, SmoothGrad, IntegratedGrad, Excitation Backpropagation, and attention models are evaluated in~\cite{Wiltschko2020,Pope2019_cvpr} with the focus on explaining the prediction of a single class.
The closest works are DeepLIFT~\cite{shrikumar17} and counterfactual explanations~\cite{Lucic2021CFGNNExplainerCE}, but they only explain the changes in a single class rather than in class distributions involving more than two classes. 

To compose an explanation, edges~\cite{gnnlrp,gnnexplainer,Faber2021kdd,shrikumar17,Lucic2021CFGNNExplainerCE} and nodes~\cite{Pope2019_cvpr} have been used to construct explanations.
Simple surrogate models~\cite{graphlime,Vu2020PGMExplainerPG} or generated samples~\cite{yuan2020xgnn} have been used.
These works cannot expose the change in information flow leading to prediction changes.

Selecting salient elements to compose an explanation is less focused.
Most work simply selects nodes or edges with the highest importance scores~\cite{gnnlrp,gnnexplainer,Faber2021kdd} with the underlying assumption that the nodes/edges contributions are independent.
Negative contributions were turned into absolute values or norms~\cite{Pope2019_cvpr}, or even be discarded~\cite{shrikumar17}, and it is not clear how negative importance scores are handled in~\cite{Wiltschko2020}.
Our convex path selection addresses both positive and negative contributions while considering the joint effect of path contributions.

Most of the prior work evaluates the faithfulness of the explanations, mainly using the closeness of the approximating prediction to the explained prediction.
To explain prediction changes, faithfulness should be evaluated based on approximation of the change.
Our metric is based on KL-divergence and designed specifically for prediction distribution changes. 

\section{Conclusions}
We studied the problem of explaining change in GNN predictions over evolving graphs.
We addressed the issues of prior works, such as irrelevant explained quantity, lack of optimality, and axiomatic attribution for GNN.
The proposed algorithm can decompose the change to information flows in the computation graphs of GNN and optimally select a small subset of paths to optimally explain the change in prediction change (using an argument of convexity).
Experimentally results showed the superiority of the proposed method over state-of-the-art baselines.
In the future work, we plan to address the situation when the edges are removed and added simultaneously.
\newpage
\bibliography{paper}
\newpage
\section{Appendix}
\subsection{Proof of Theorem 1}
\begin{proof}
Considering the path $p=(I,\dots,U,V,\dots J)$ on the graph $G_1$, for the DeepLIFT on the graph, we let the reference activations be  set to the empty graph. Then, the
reference activation of $p[t]$ is zero. Thus, $\Delta h_{p[t]}^{(t)}=h^{(t)}_{p[t]}(G_1), \Delta z_{p[t]}^{(t)}=z^{(t)}_{p[t]}(G_1), m_{\Delta h^{(t-1)}_u \Delta h^{(t)}_v}=\frac{\Delta h^{(t)}_v}{\Delta z^{(t)}_v } \times \theta_{u,v}^{(t)}$.
While, for the GNN-LRP method when $\gamma =0$, on the message passing layers, $R_j=z_j$, we note $\textnormal{LRP}_{u,v}^{(t)}=\frac{h_{u}^{(t-1)}\theta^{(t)}_{u,v}}{\sum \limits_{U \in N(V)} \sum \limits_{u} h_{u}^{(t-1)}\theta^{(t)}_{u,v}}=\frac{h_{u}^{(t-1)}\theta^{(t)}_{u,v}}{z_{v}^{(t)}}$ that represents the allocation rule of neuron $v$ to its predecessor neuron $u$ in the GNN-LRP method. The contribution of this path is 
\begin{eqnarray}
    R_p &=&\sum_{i} \dots  \sum_{p[T-1]} \textnormal{LRP}_{i,p[1]}^{(1)}  \ldots \textnormal{LRP}_{p[T-1],j}^{(T)} R_j \nonumber\\
    ~&=&\sum_{i} \dots  \sum_{p[T-1]} \frac{h_{i}^{(0)}\theta^{(1)}_{i,p[1]}}{z_{p[1]}^{(1)}} \dots \frac{h_{p[T-1]}^{(T-1)}\theta^{(T)}_{p[T-1],j}}{z_{j}} z_{j} \nonumber \\
    ~&=&\sum_{i}h_i^{(0)}\sum_{p[1]}\dots\sum_{p[T-1]}\frac{h_{p[1]}^{(1)}\theta^{(1)}_{i,p[1]}}{z_{p[1]}^{(1)}}\dots\theta^{(T)}_{p[T-1],j} \nonumber \\
    ~&=&\sum_{i} m_{\Delta h_i^{(0)} \Delta z_j} h_i^{(0)} \nonumber \\
    ~&=& C_{p,j} \nonumber
\end{eqnarray}

\end{proof}

\subsection{Datasets}
\begin{itemize}[leftmargin=*]
  \item 
Citeseer, Cora, and PubMed~\cite{kipf2017_iclr}:
each node is a paper with a bag-of-words feature vector, and nodes are connected by the citation relationship.
The goal is to predict the research area of each paper.
\item
Amazon-Computer (Amazon-C) and Amazon-Photo (Amazon-P)~\cite{shchur2018pitfalls}: segments of the Amazon co-purchase graph,
where nodes represent products and edges indicate that two products are frequently purchased together,
node features are the bag-of-words vectors of the product reviews.
\item
Coauthor-Computer and Coauthor-Physics: co-authorship graphs based on the Microsoft Academic Graph from the KDD Cup 2016 Challenge.
We represent authors as nodes, that are connected by an edge if they co-authored a paper~\cite{shchur2018pitfalls}.
Node features represent paper keywords for each author’s papers.
\end{itemize}

\subsection{Experimental setup}
We choose element-wise sum as the $f_\textnormal{AGG}$ function to ensure that the attribution by DeepLIFT can preserve  the  total  change  in  the  logits. We train the GNN model with two layers and three layers on seven datasets and fix these model parameters. 200 edges randomly selected are added to simulate the graph evolution. To show that as $n$ increases,  $\pr_J(G_n)$ is gradually approaching $\pr_J(G_1)$, we let n gradually increase. Moreover, $m$ is different for the different nodes in $V^{*}$, so we choose $n$ according to $m$. If $10<m \leq 30,n=[1,2,3,4,5,6,7,8,9,10]$, if $30<m \leq 100,n=[10,12,14,16,18,20,22,24,26,28]$ and if $m \geq 100 ,n=[10,15,20,25,30,35,40,45,50,55].$

\subsection{Standard deviation for $\mathbf{Fidelity}_{\textnormal{KL}}^{-}$}
\begin{figure}[htbp]
  \subfloat{\includegraphics[width = 0.25\textwidth]{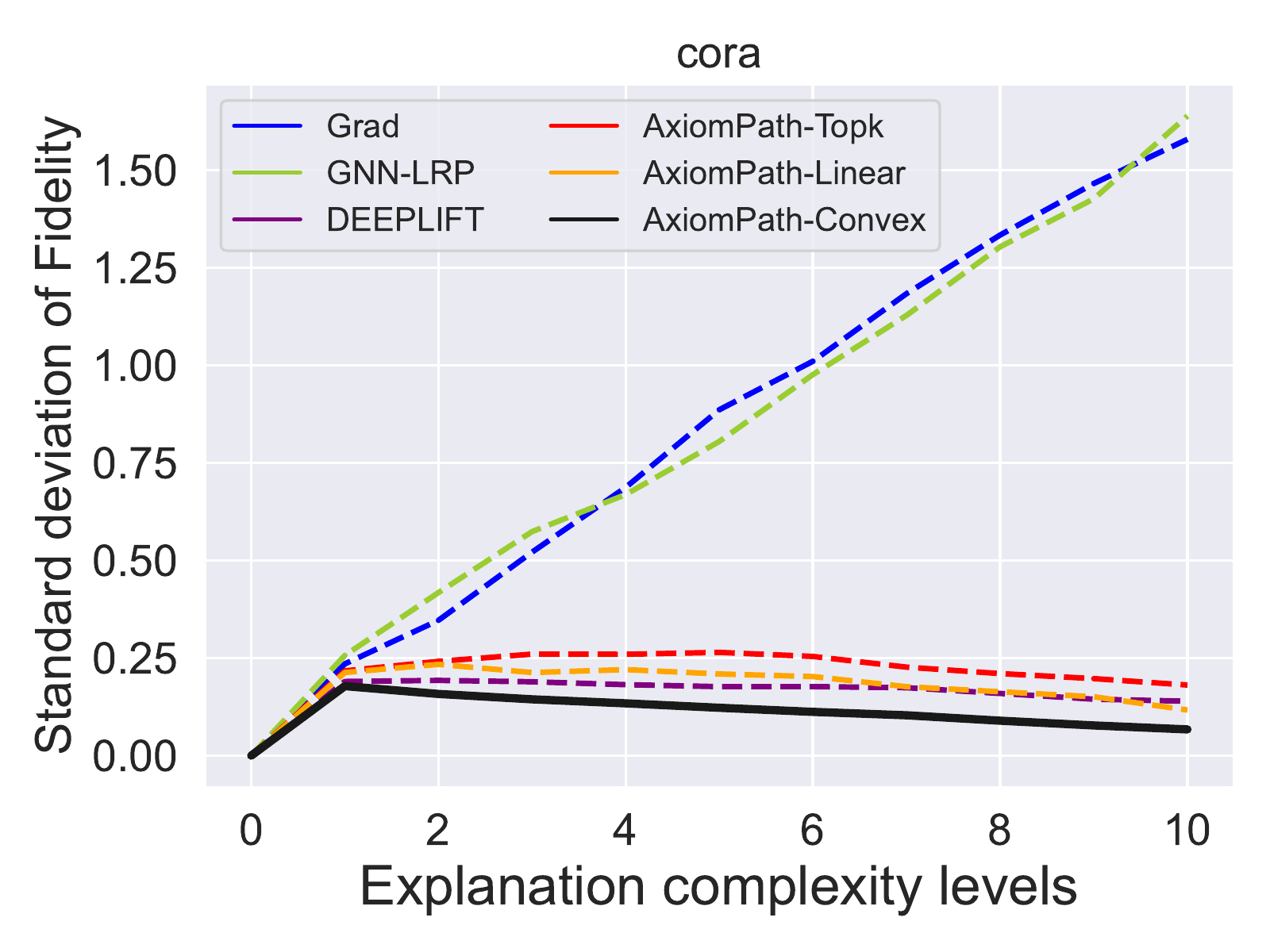}
  }
  \subfloat{\includegraphics[width = 0.25\textwidth]{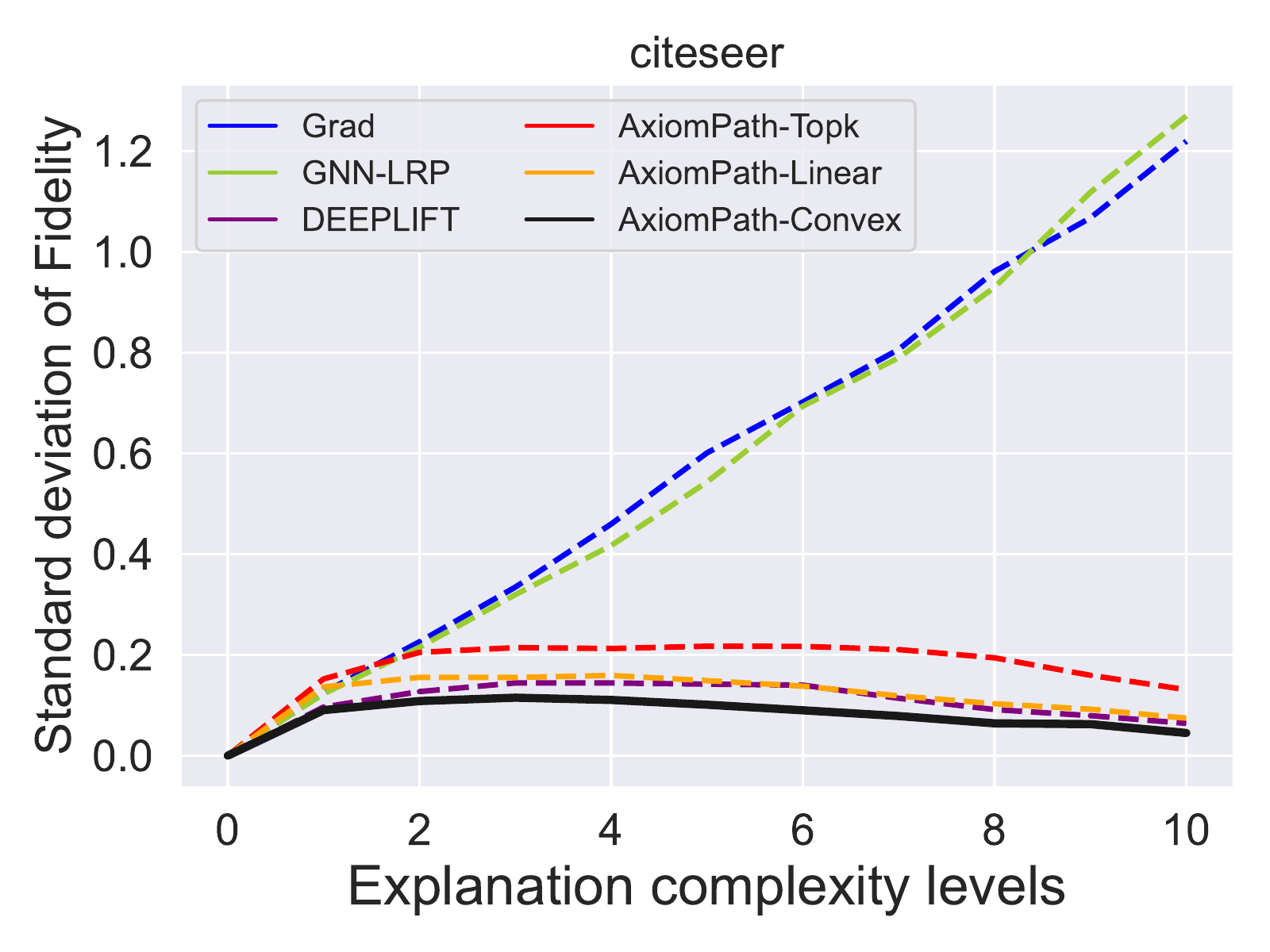}
  }\\
  \subfloat{\includegraphics[width = 0.25\textwidth]{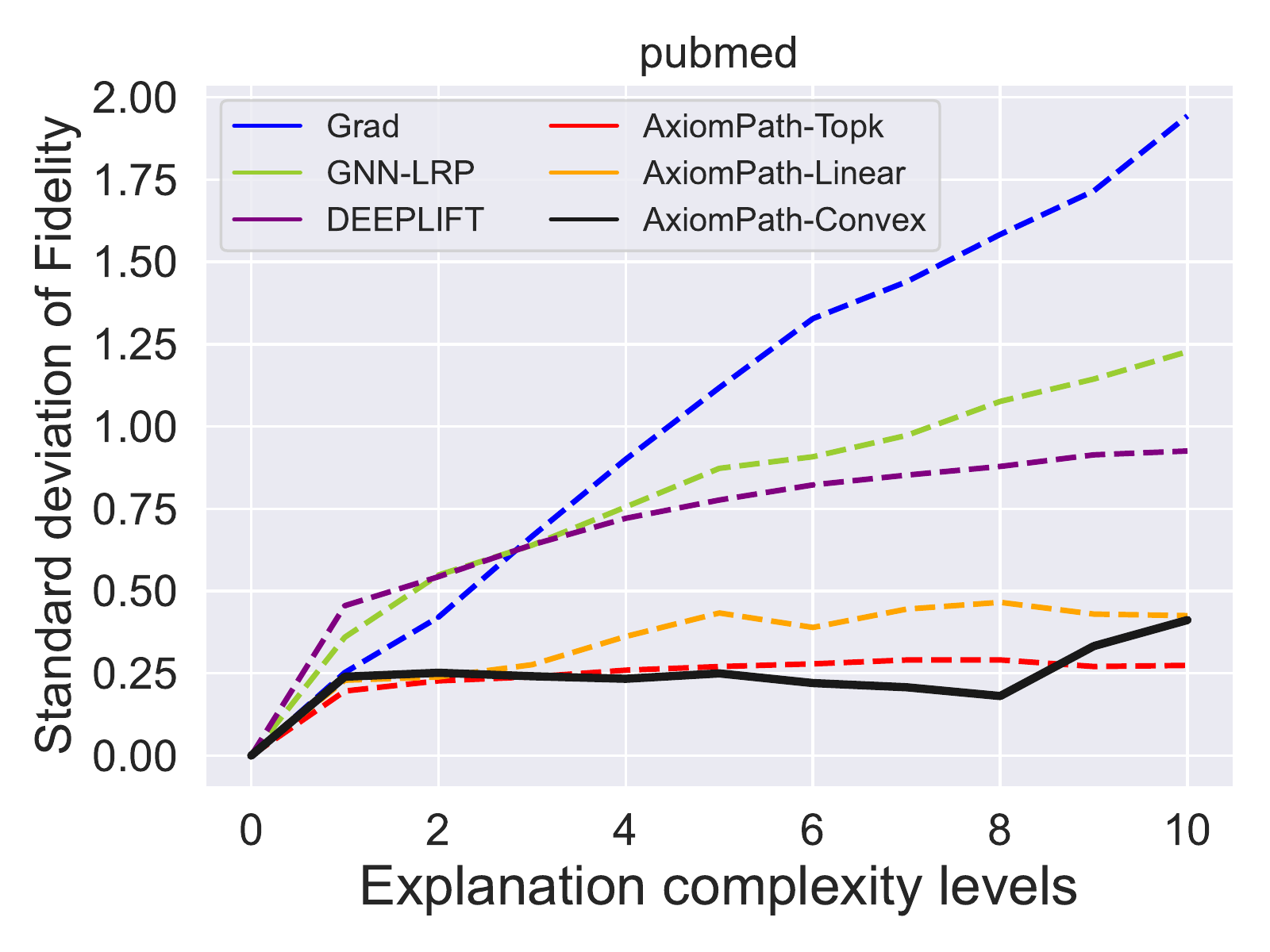}
  }
  \subfloat{\includegraphics[width = 0.25\textwidth]{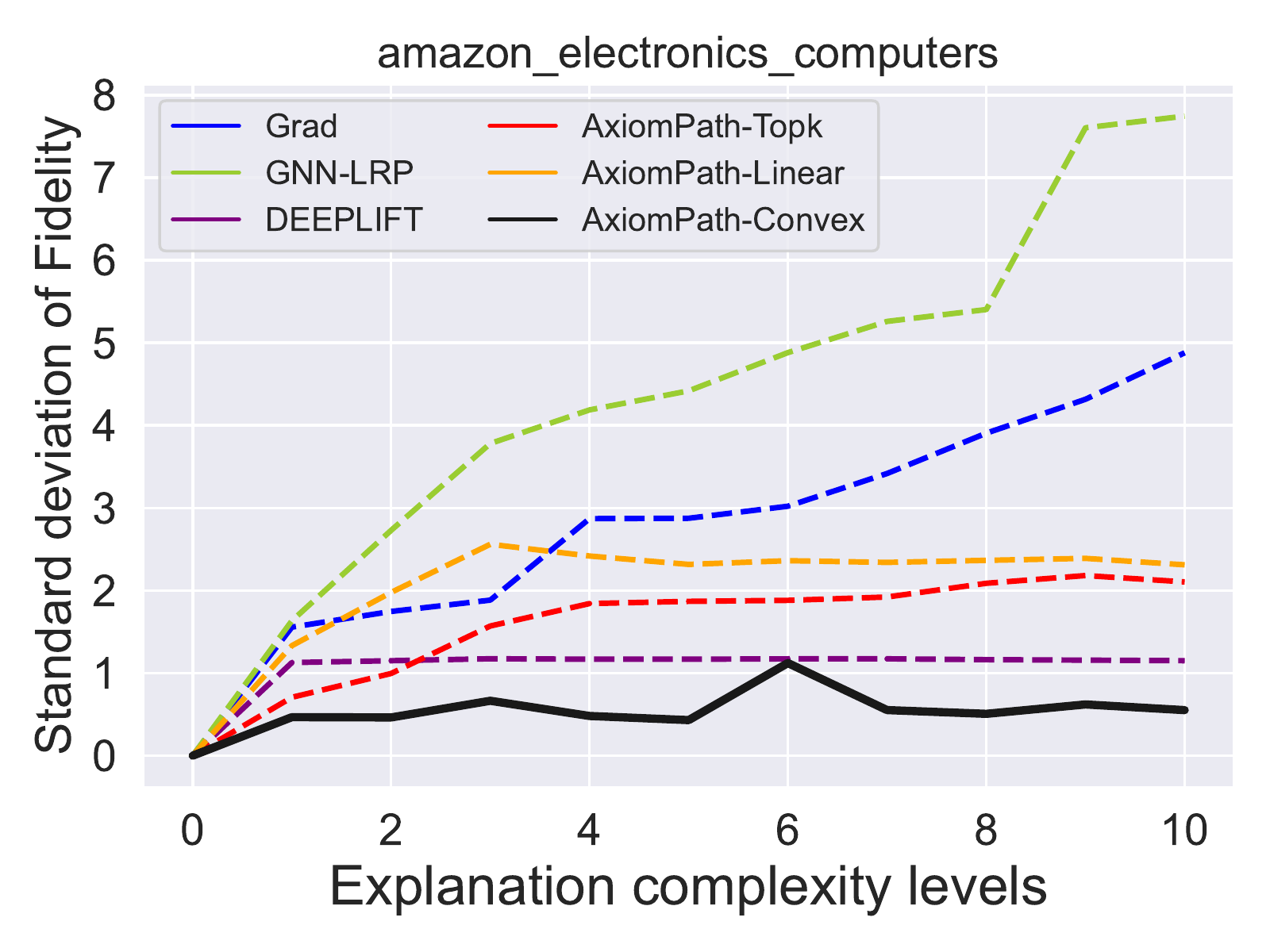}
  }\\
  \subfloat{\includegraphics[width = 0.25\textwidth]{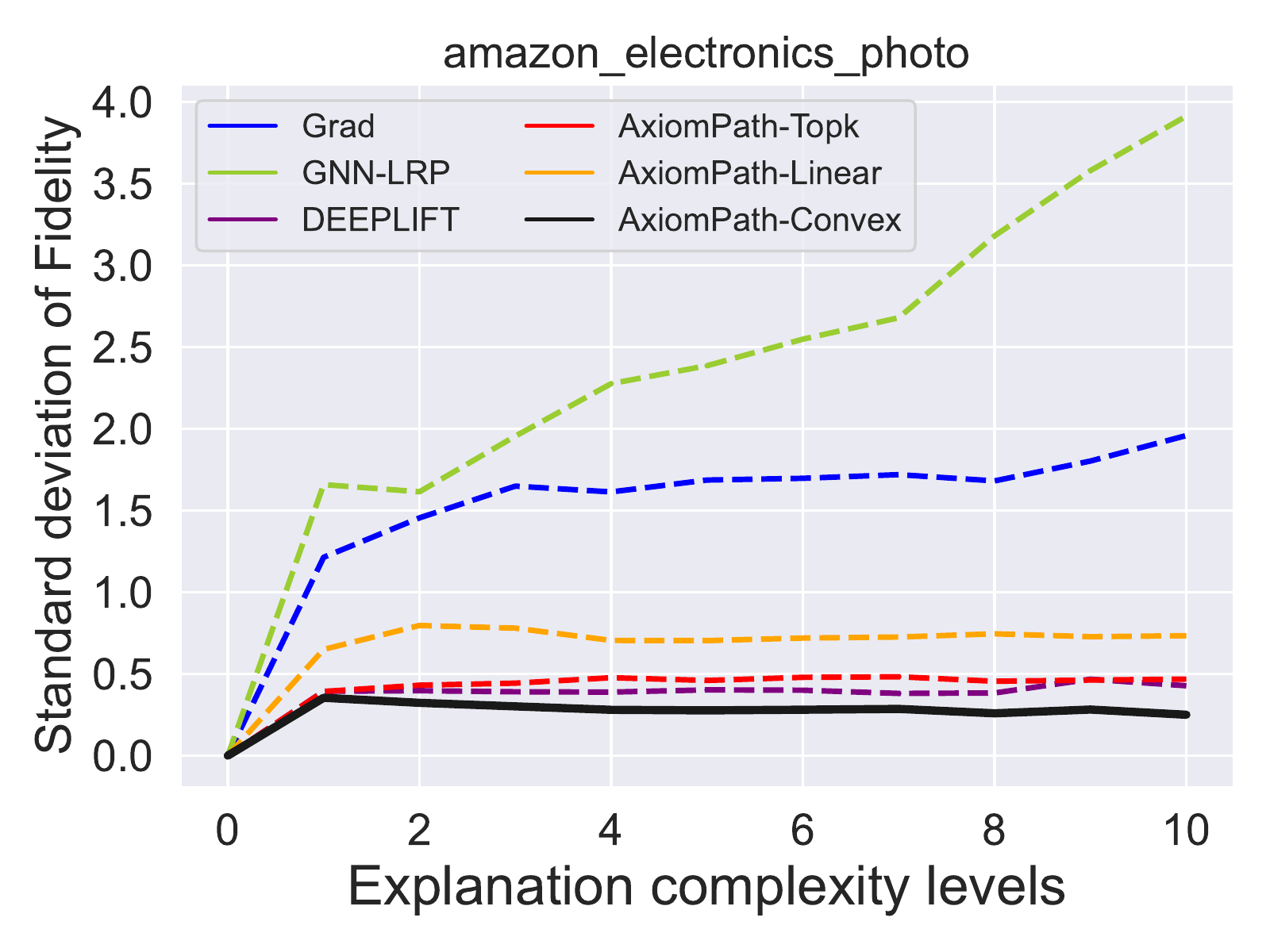}
  }
  \subfloat{\includegraphics[width = 0.25\textwidth]{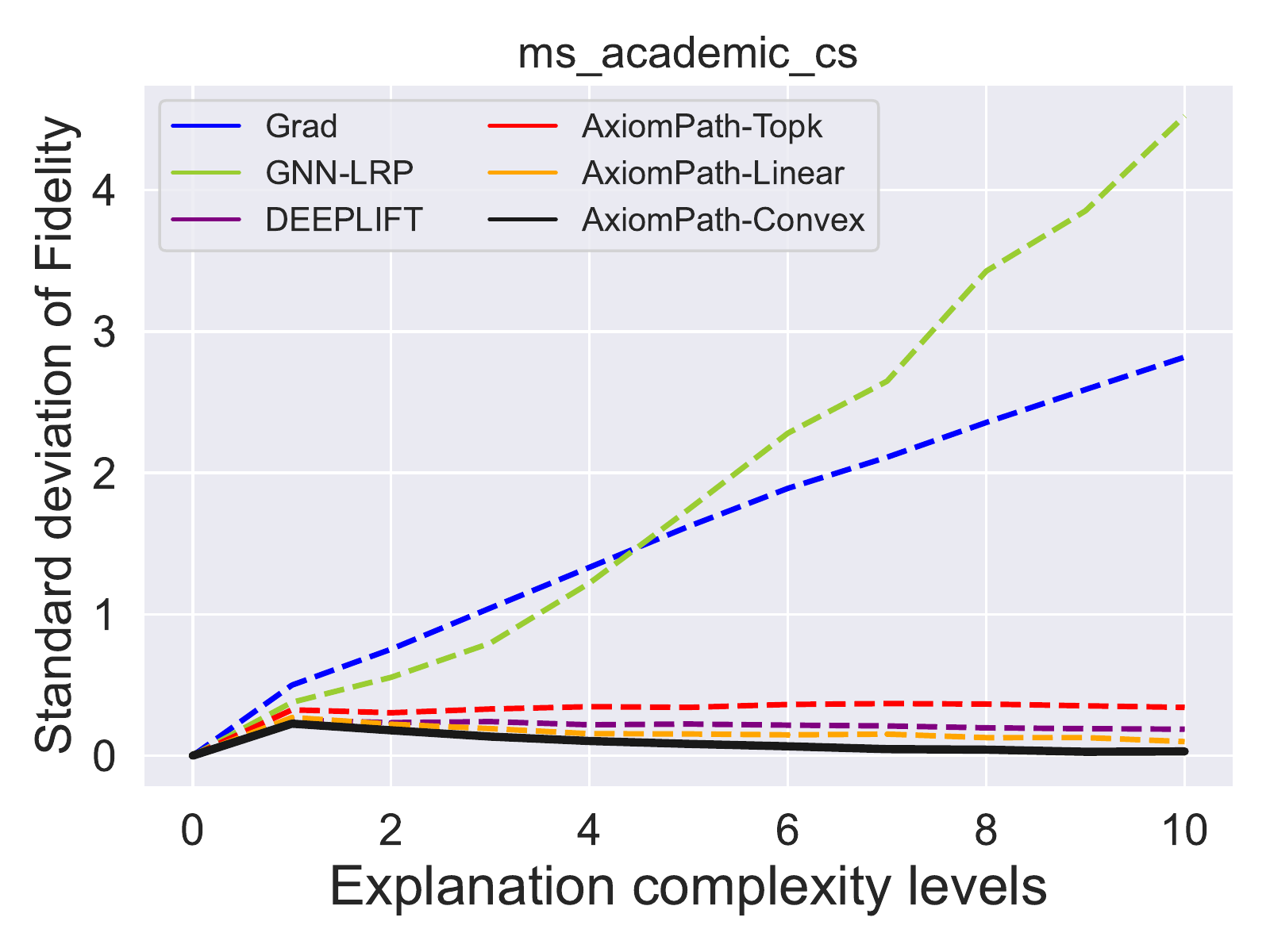}
  }\\
  \subfloat{\includegraphics[width = 0.25\textwidth]{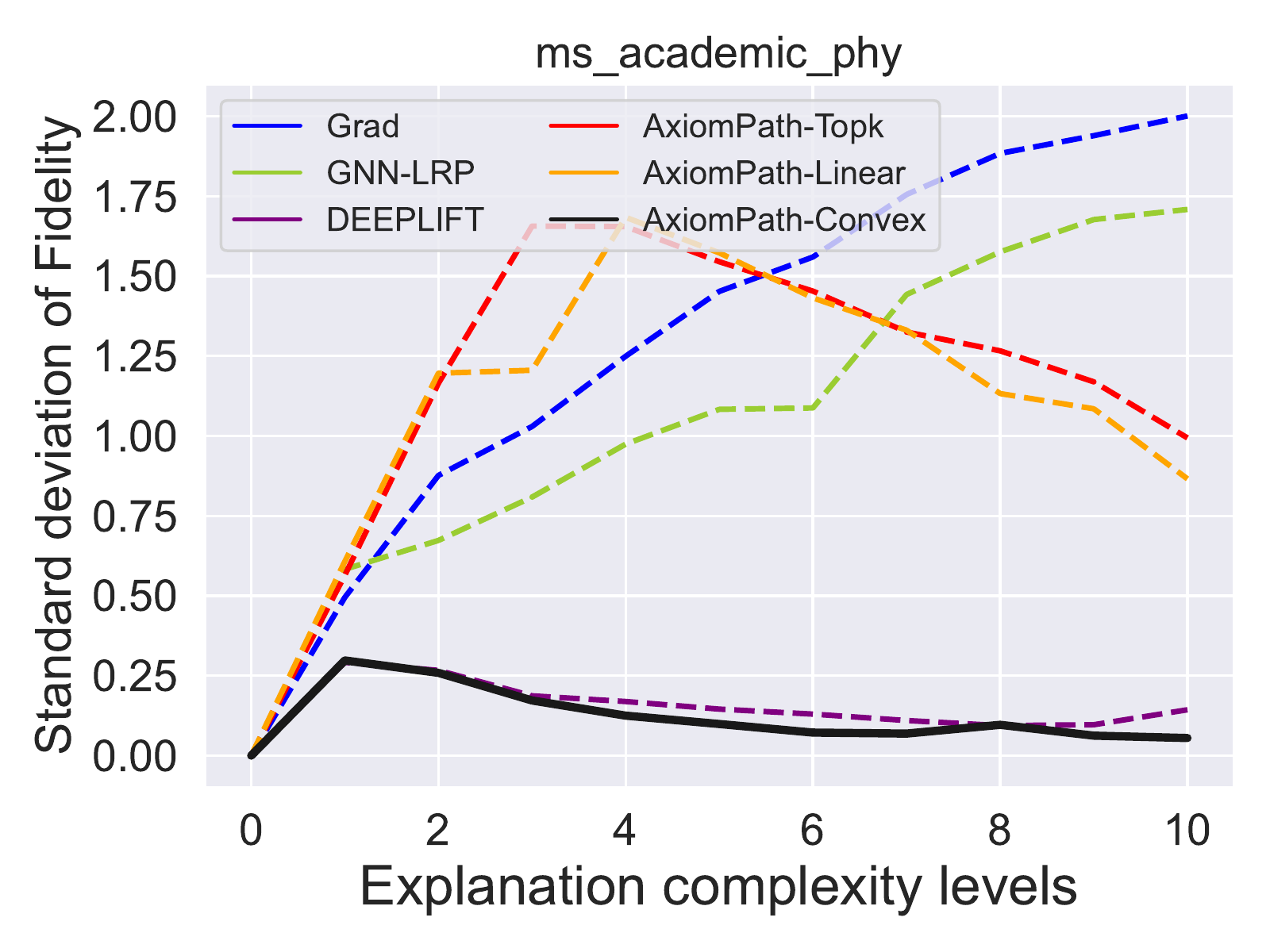}
  }
  \caption{\small Standard deviation of \textbf{Fidelity}$_\textnormal{KL}^{-}$ over all datasets.
  AxiomPath-Convex has the best performance.
  }
\label{fig:overall_std}
\end{figure}

From Figure~\ref{fig:overall_std}, we can see that although on the PubMed dataset, the standard deviation of AxiomPath-Convex is higher than the standard deviation of AxiomPath-Topk when the Explanation complexity level is 9 or 10, AxiomPath-Convex has the smallest standard deviation over all levels of explanation complexities on six datasets
(Cora, Citeseer, Amazon-C, Amazon-P,  Coauthor-Computer and Coauthor-Physics). Thus, AxiomPath-Convex has the best performance.
\subsection{Performance evaluation and comparison}
In the main texts, we show the mean \textbf{Fidelity}$_\textnormal{KL}^{-}$ over all datasets and the number of layers of the fixed GNN model is two. When the number of GNN model layers is three, we also run the evolution and explanation for ten times to calculate the means and standard deviations of the \textbf{Fidelity}$_\textnormal{KL}^{-}$ over
target nodes.
\begin{figure}[htbp]
  \subfloat{\includegraphics[width = 0.25\textwidth]{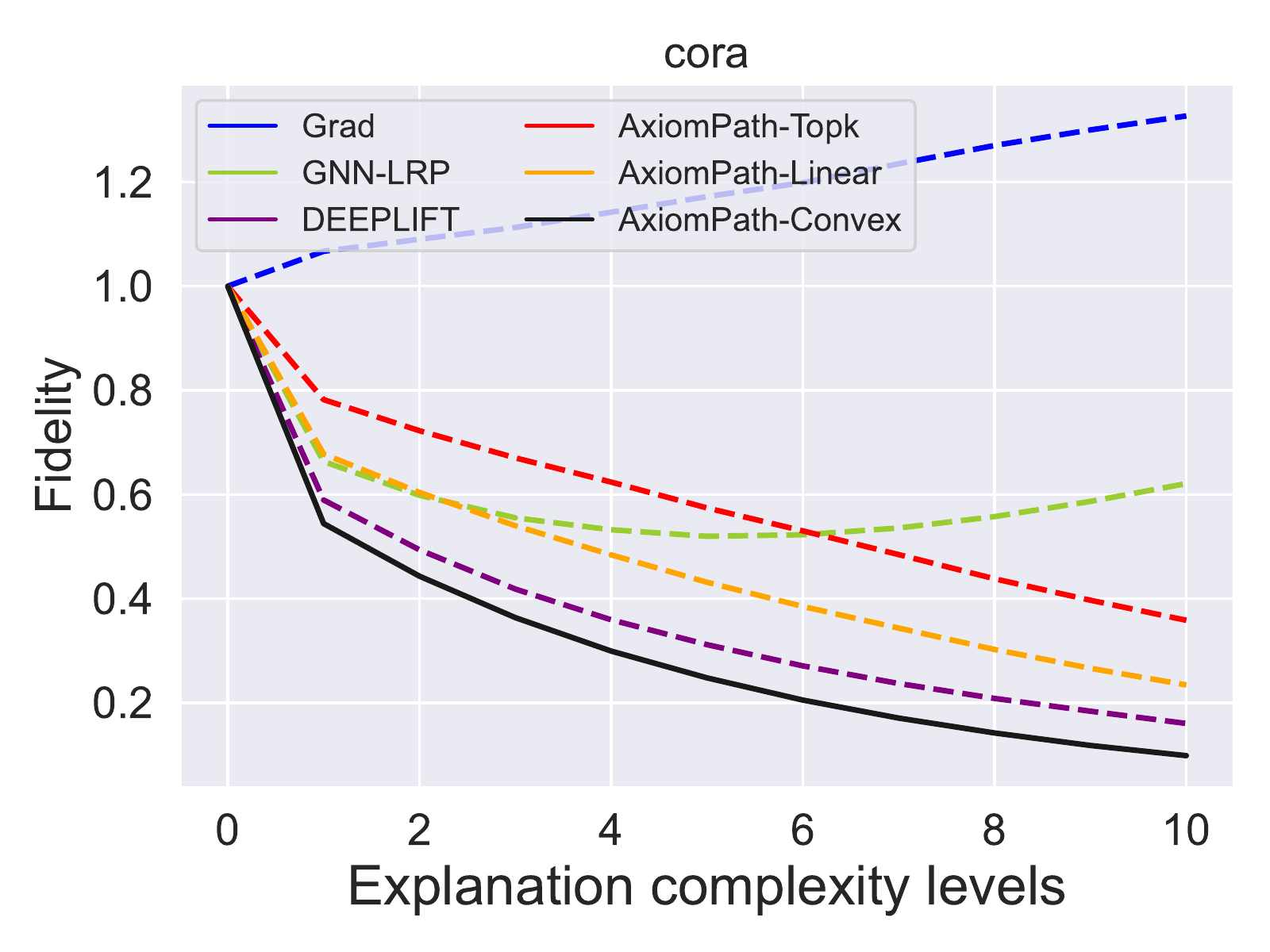}
  }
  \subfloat{\includegraphics[width = 0.25\textwidth]{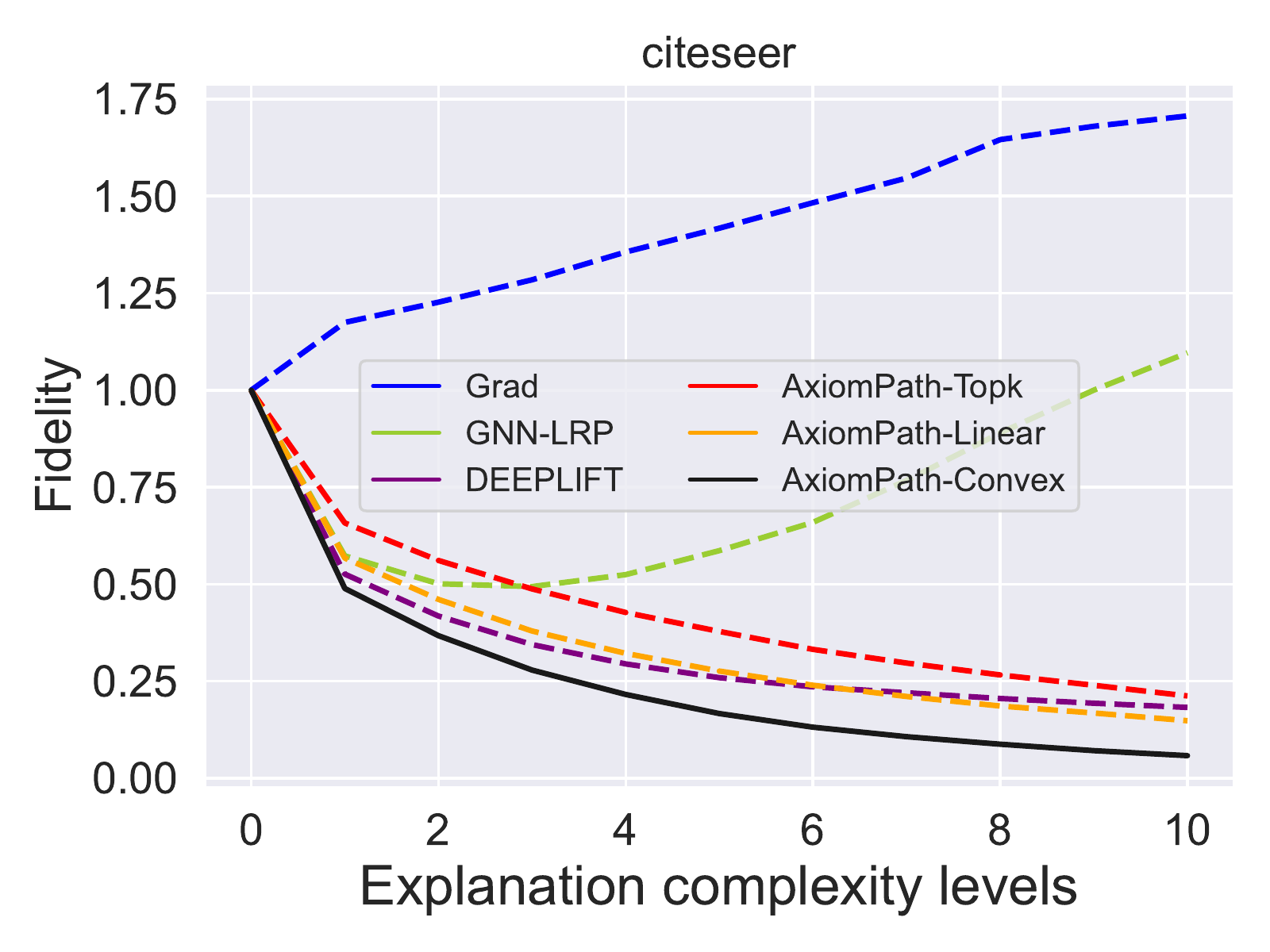}
  }\\
  \subfloat{\includegraphics[width = 0.25\textwidth]{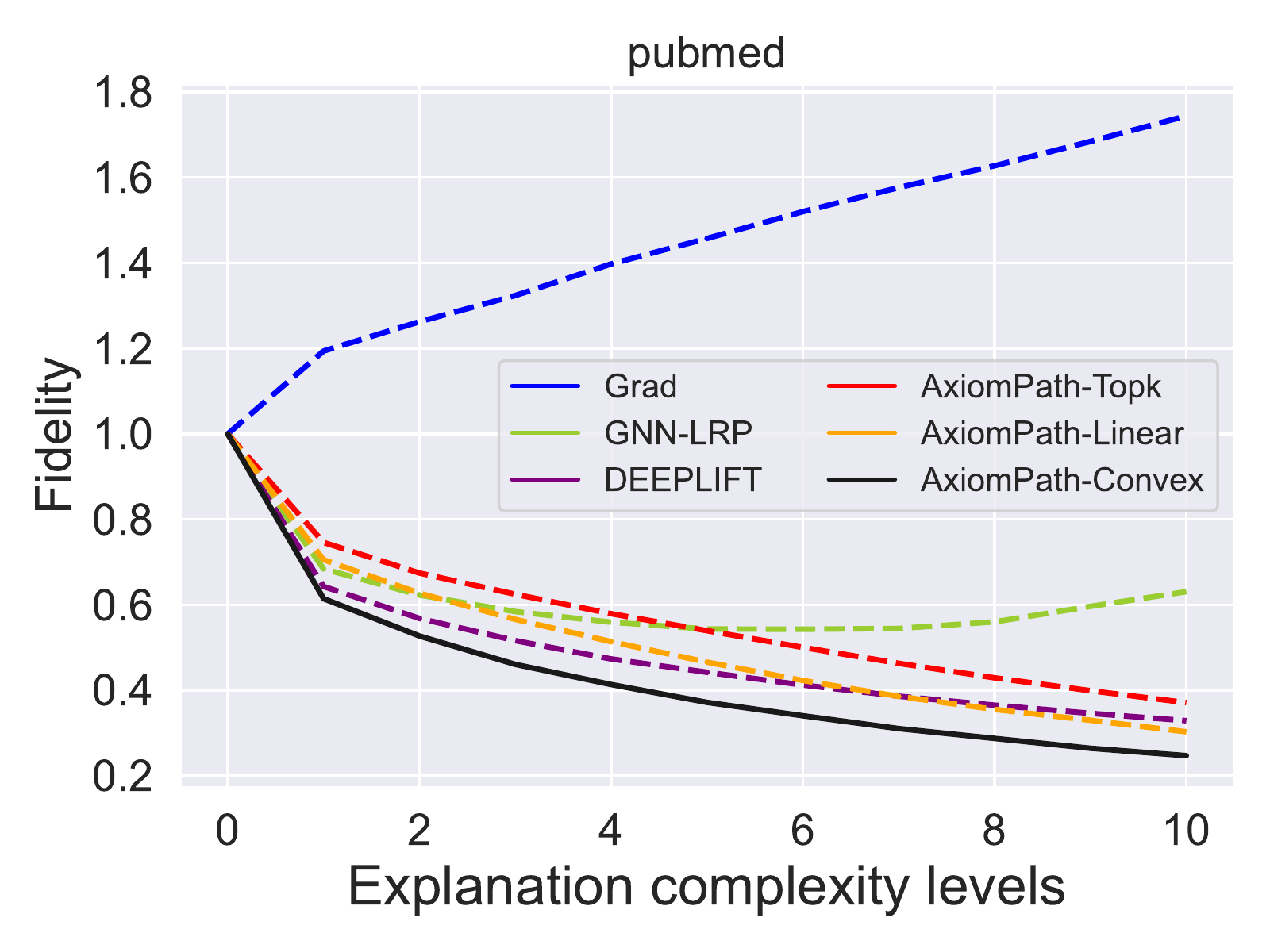}
  }
  \subfloat{\includegraphics[width = 0.25\textwidth]{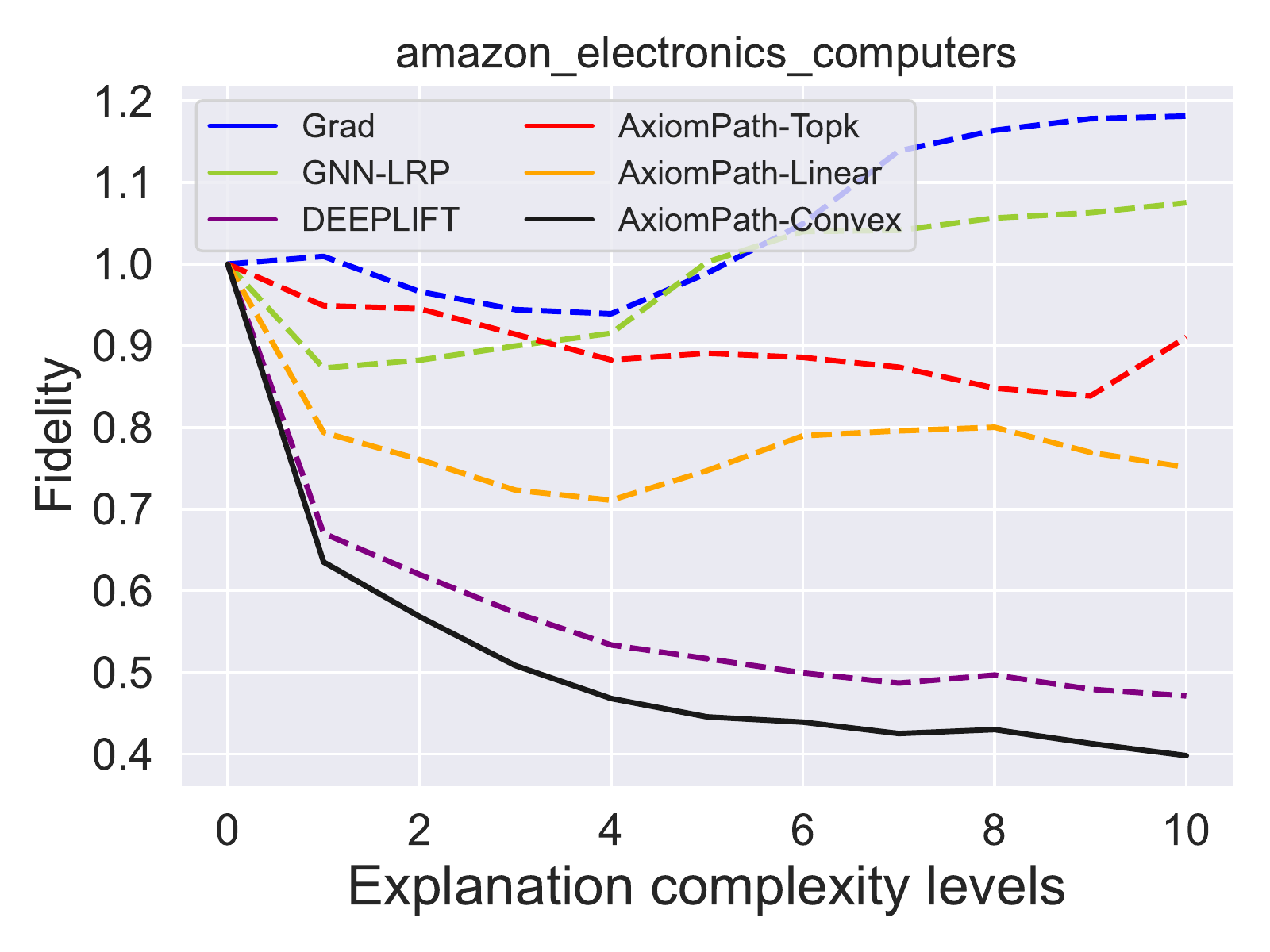}
  }\\
  \subfloat{\includegraphics[width = 0.25\textwidth]{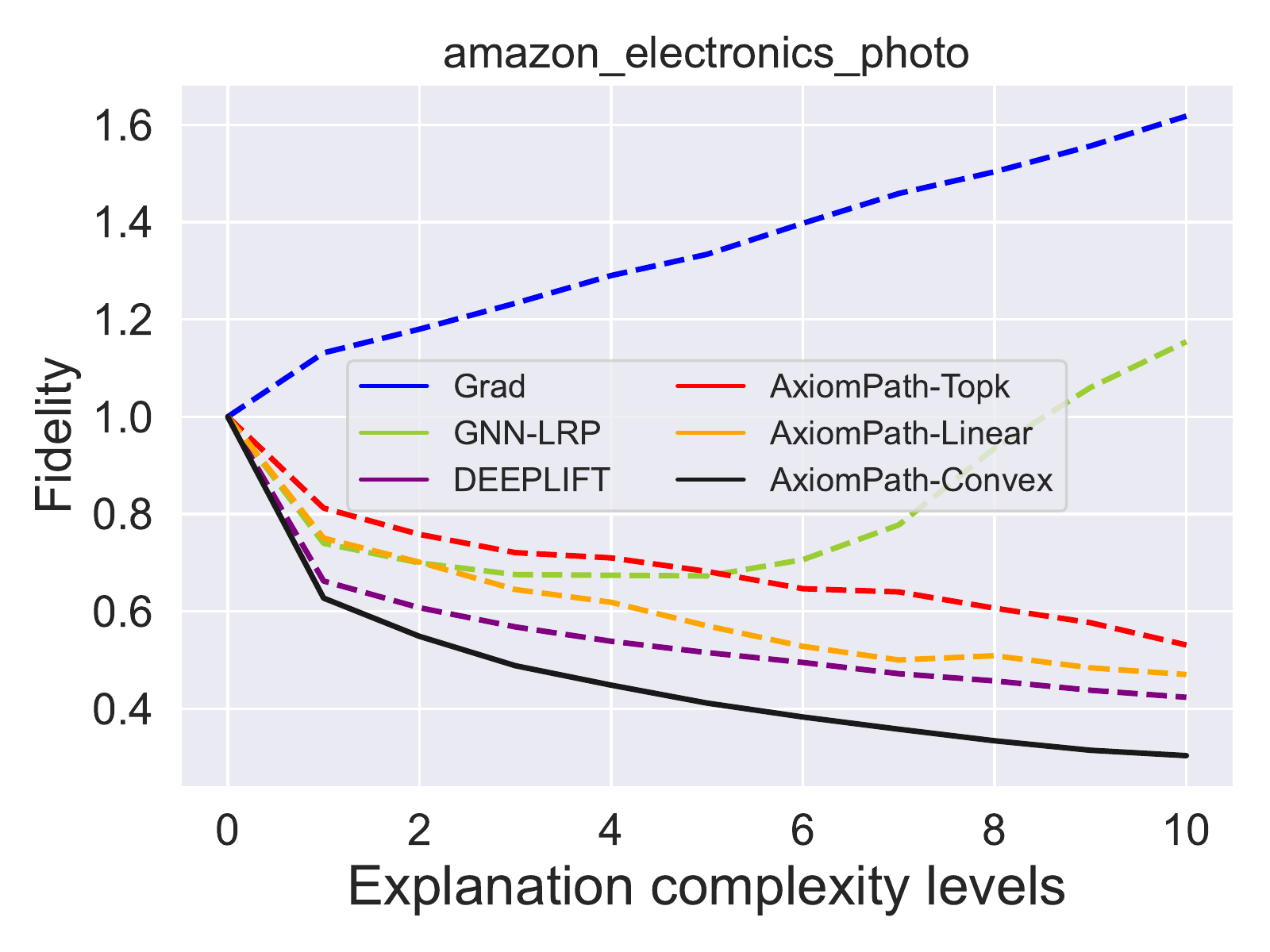}
  }
  \subfloat{\includegraphics[width = 0.25\textwidth]{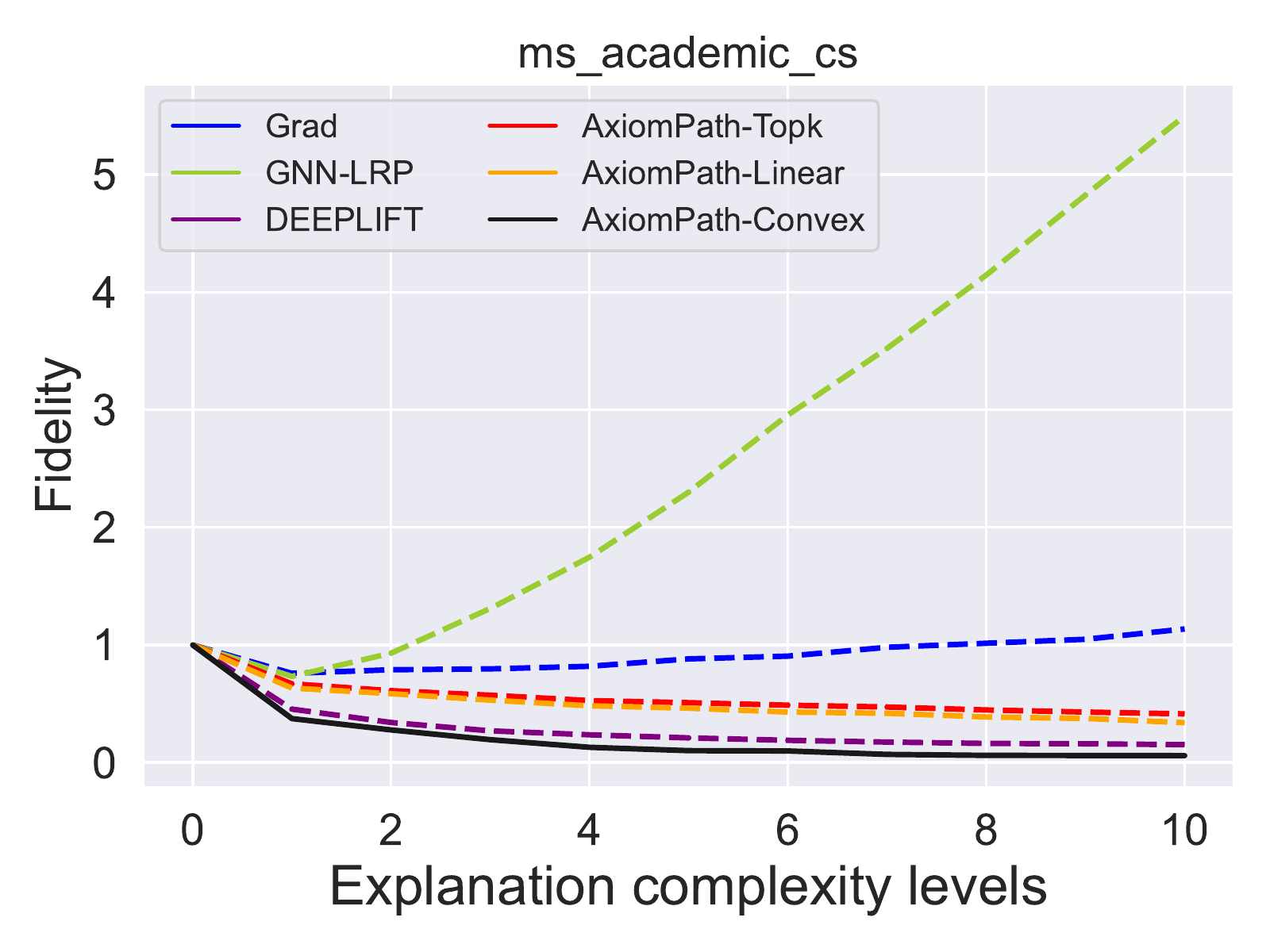}
  }\\
  \subfloat{\includegraphics[width = 0.25\textwidth]{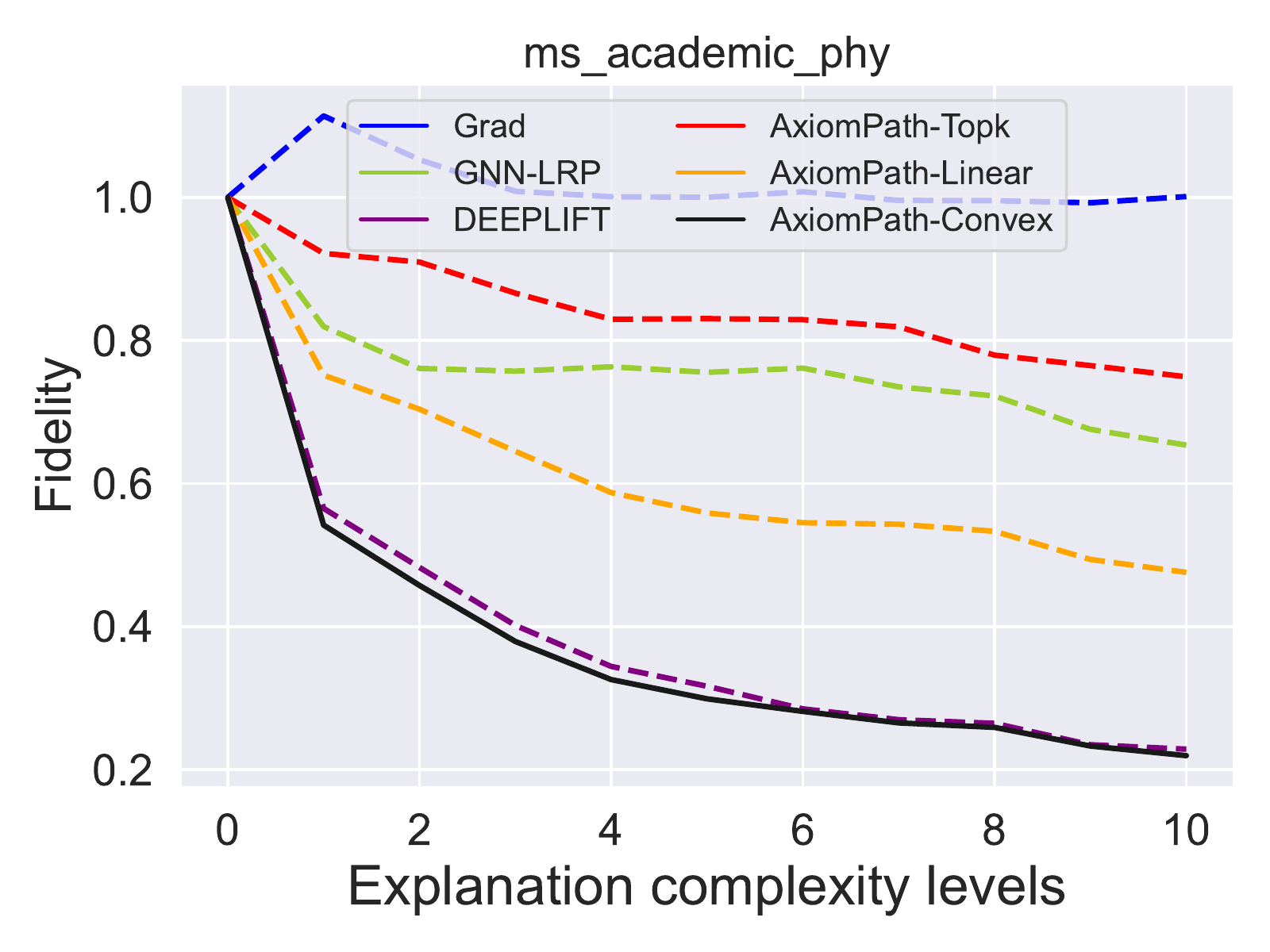}
  }
  \caption{\small Mean  \textbf{Fidelity}$_\textnormal{KL}^{-}$ over all datasets when the number of layers of the fixed GNN model is three. AxiomPath-Convex has the best performance.
  }
\label{fig:overall_layer3_kl}
\end{figure}

\begin{figure}[htb]
  \subfloat{\includegraphics[width = 0.25\textwidth]{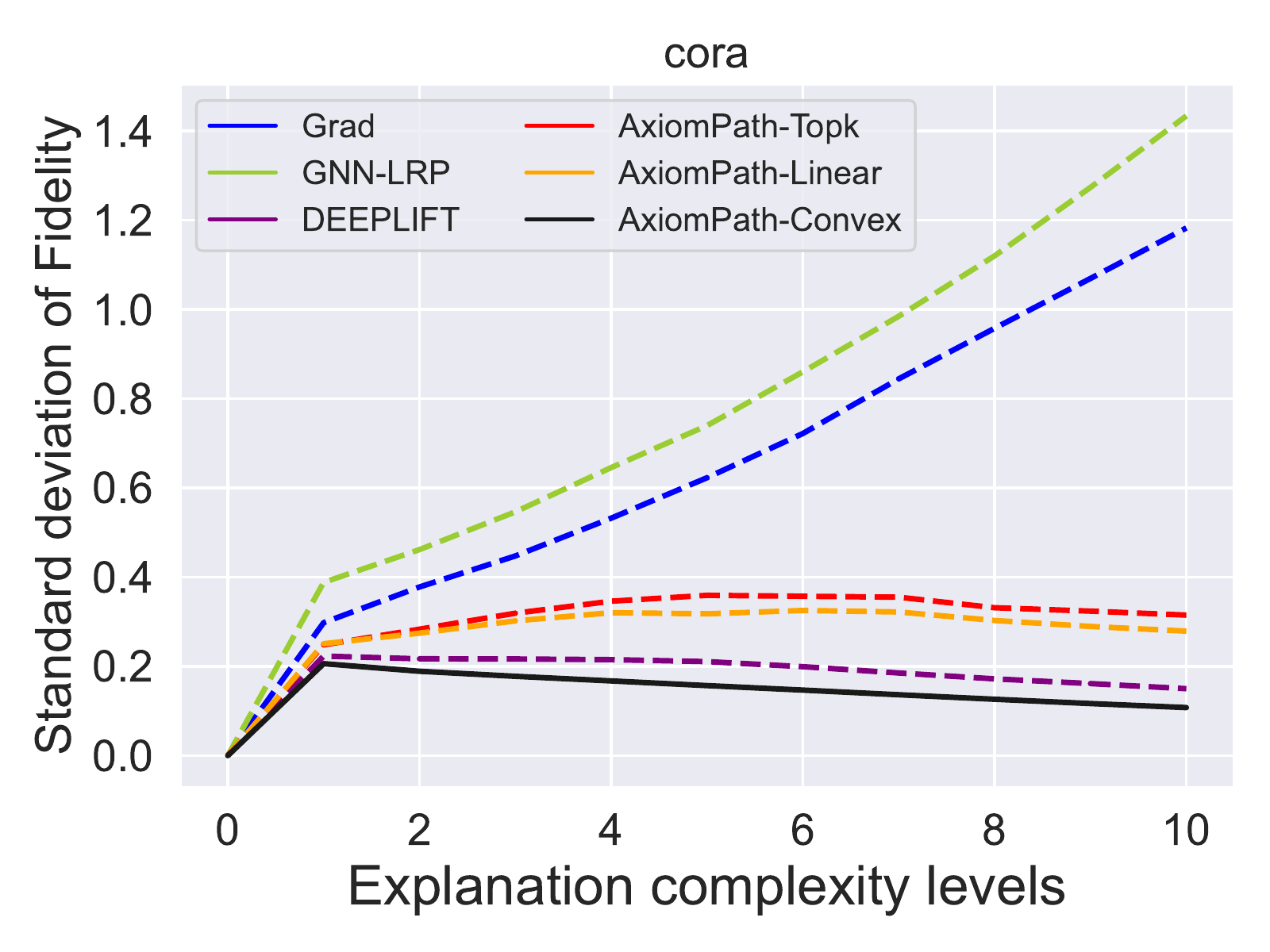}
  }
  \subfloat{\includegraphics[width = 0.25\textwidth]{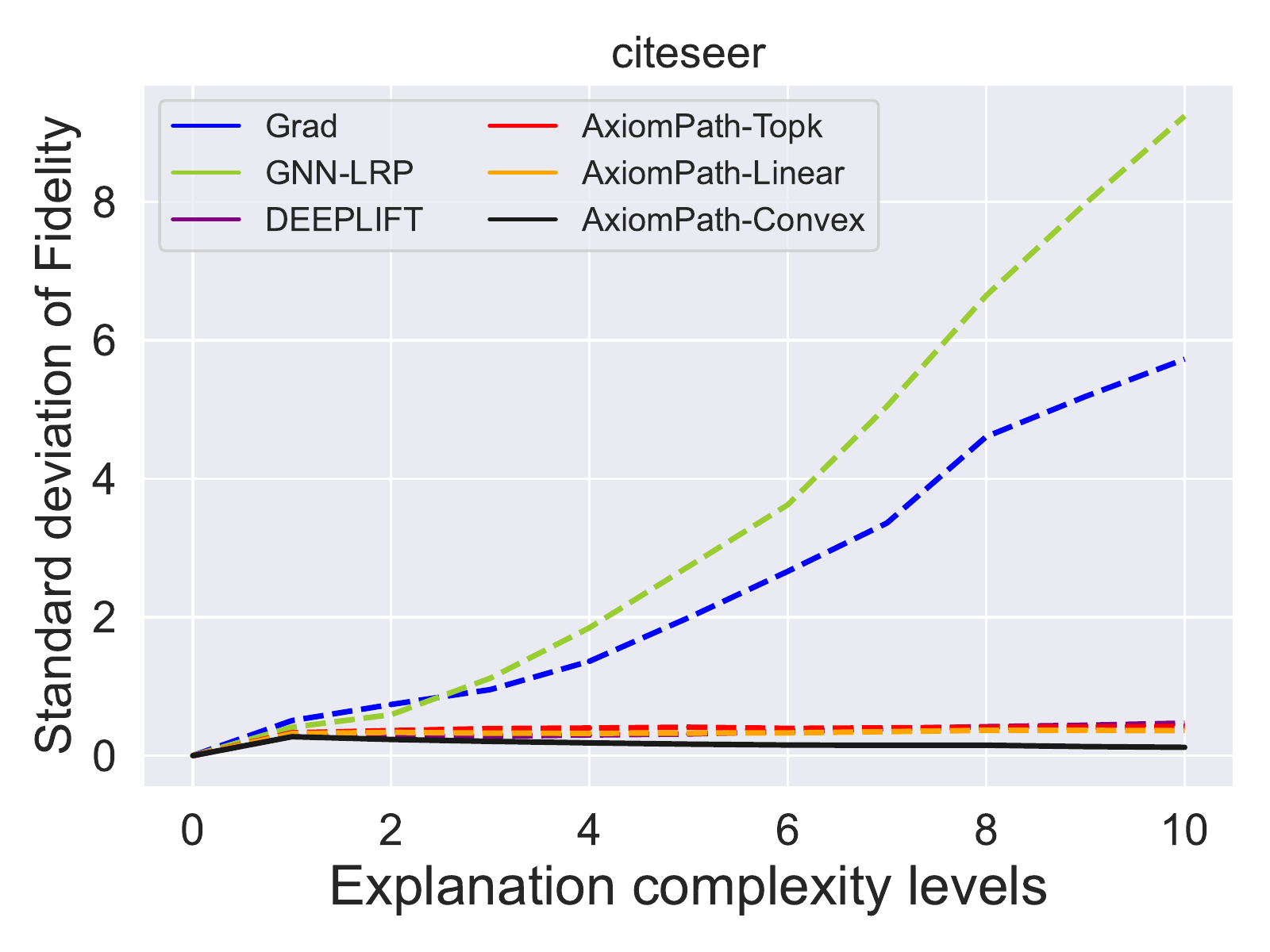}
  }\\
  \subfloat{\includegraphics[width = 0.25\textwidth]{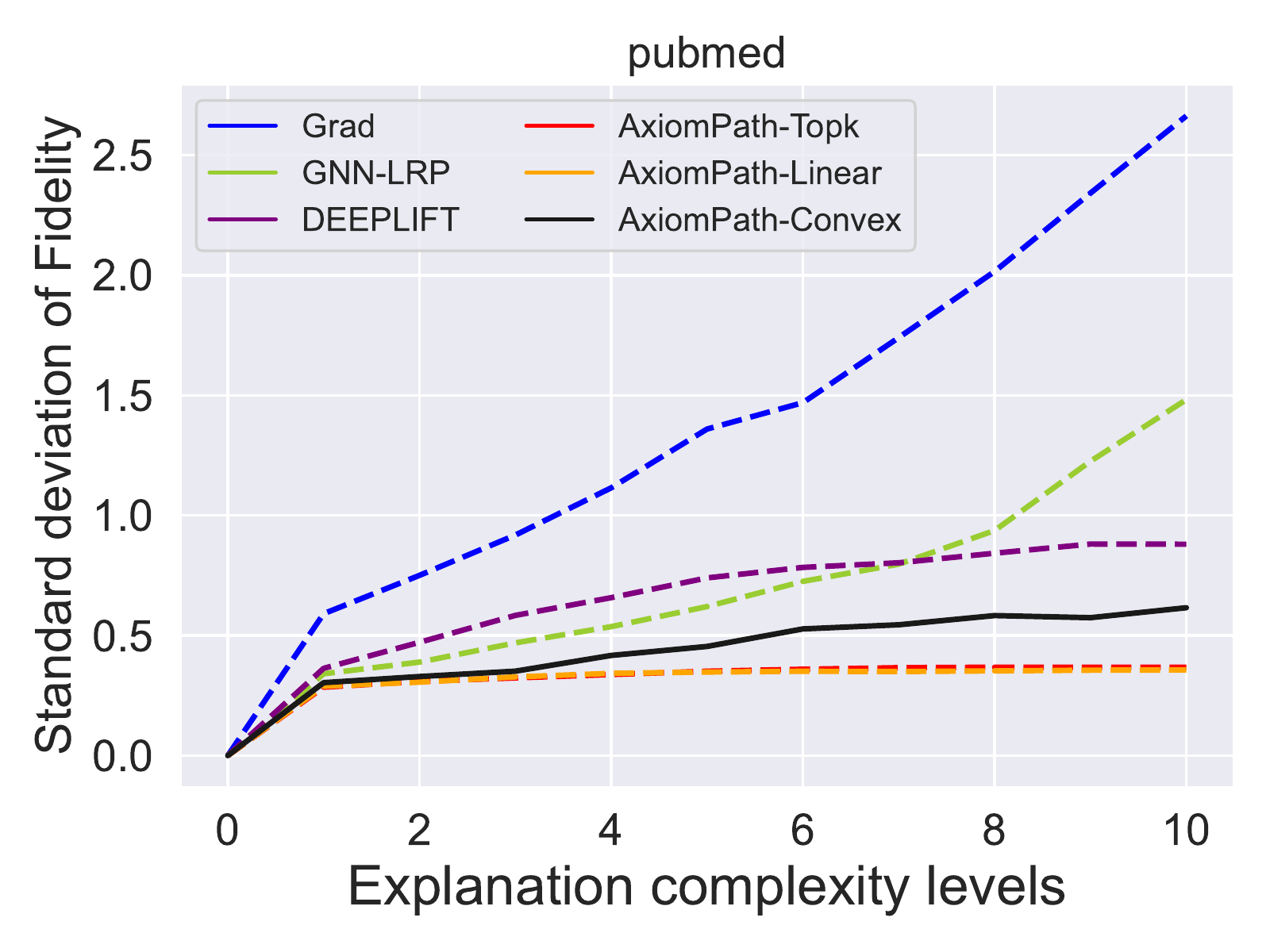}
  }
  \subfloat{\includegraphics[width = 0.25\textwidth]{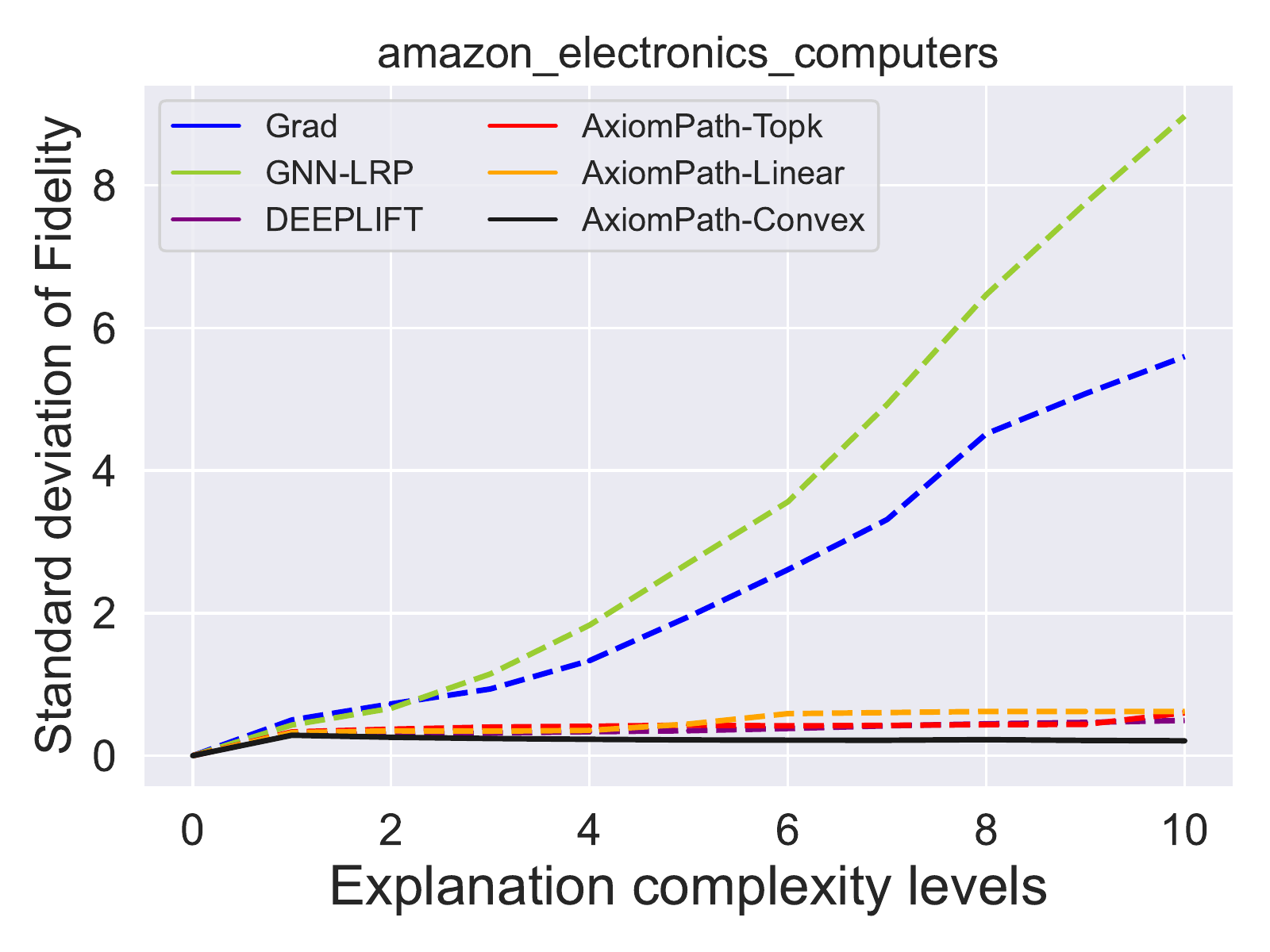}
  }\\
  \subfloat{\includegraphics[width = 0.25\textwidth]{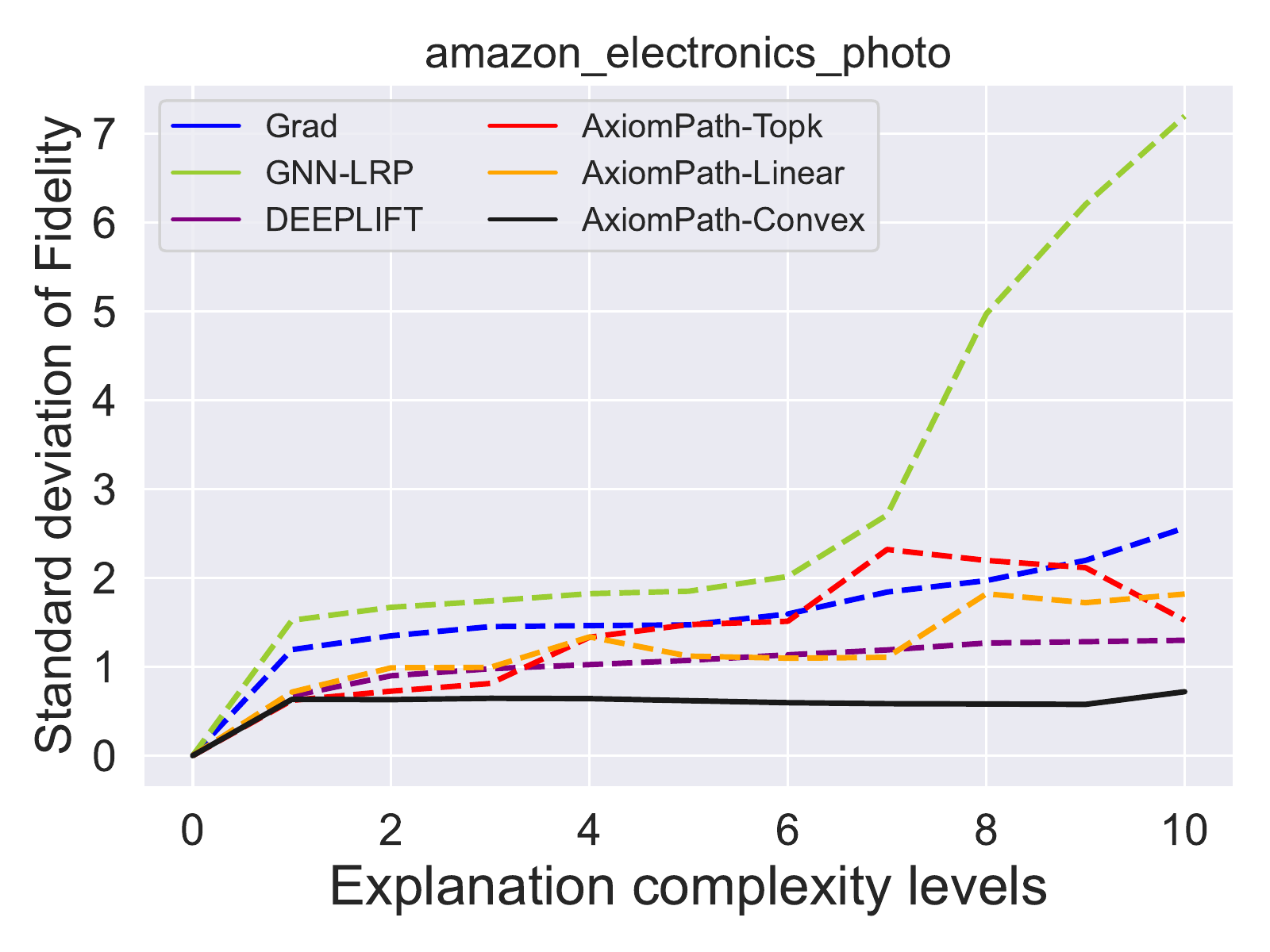}
  }
  \subfloat{\includegraphics[width = 0.25\textwidth]{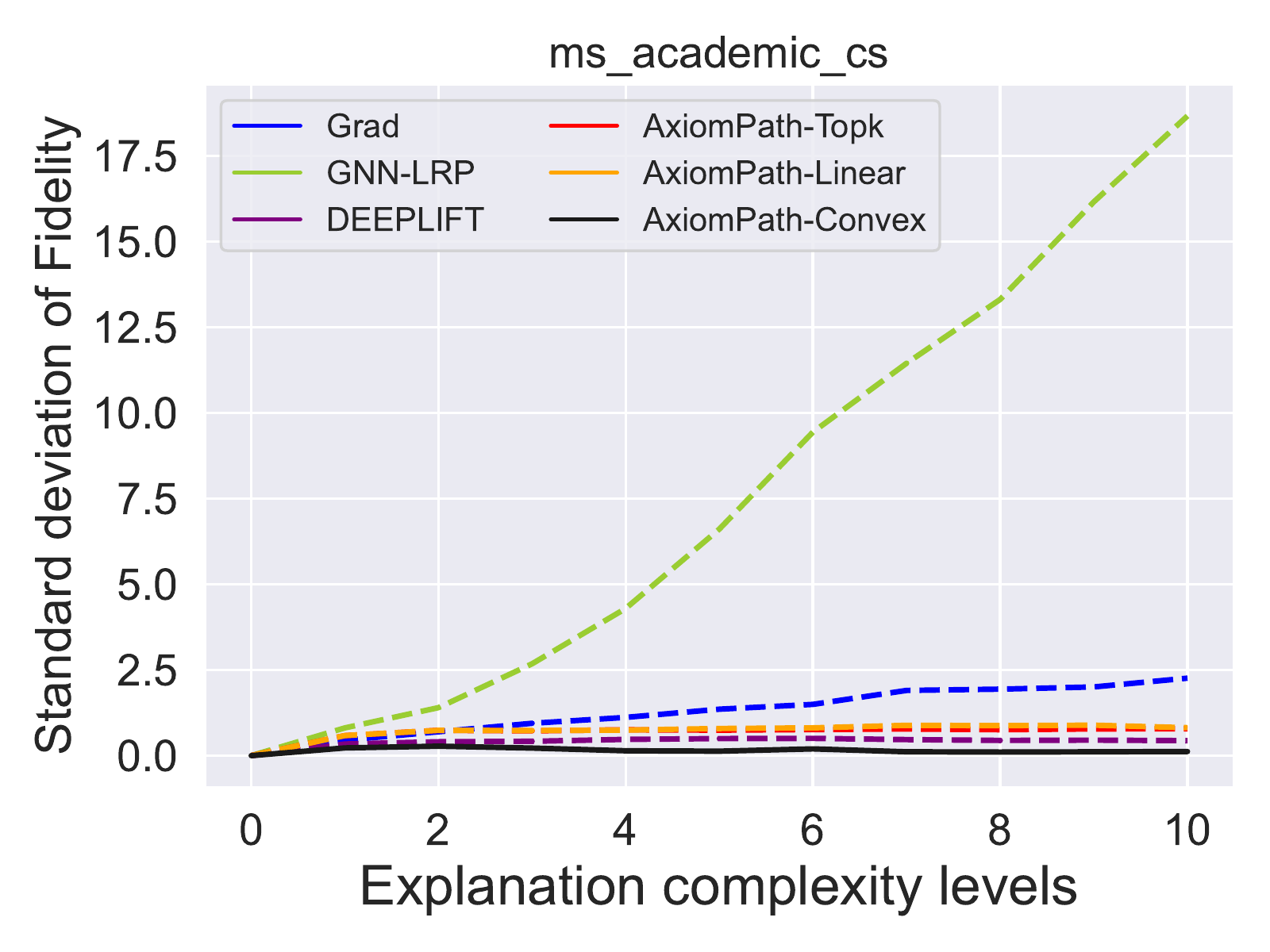}
  }\\
  \subfloat{\includegraphics[width = 0.25\textwidth]{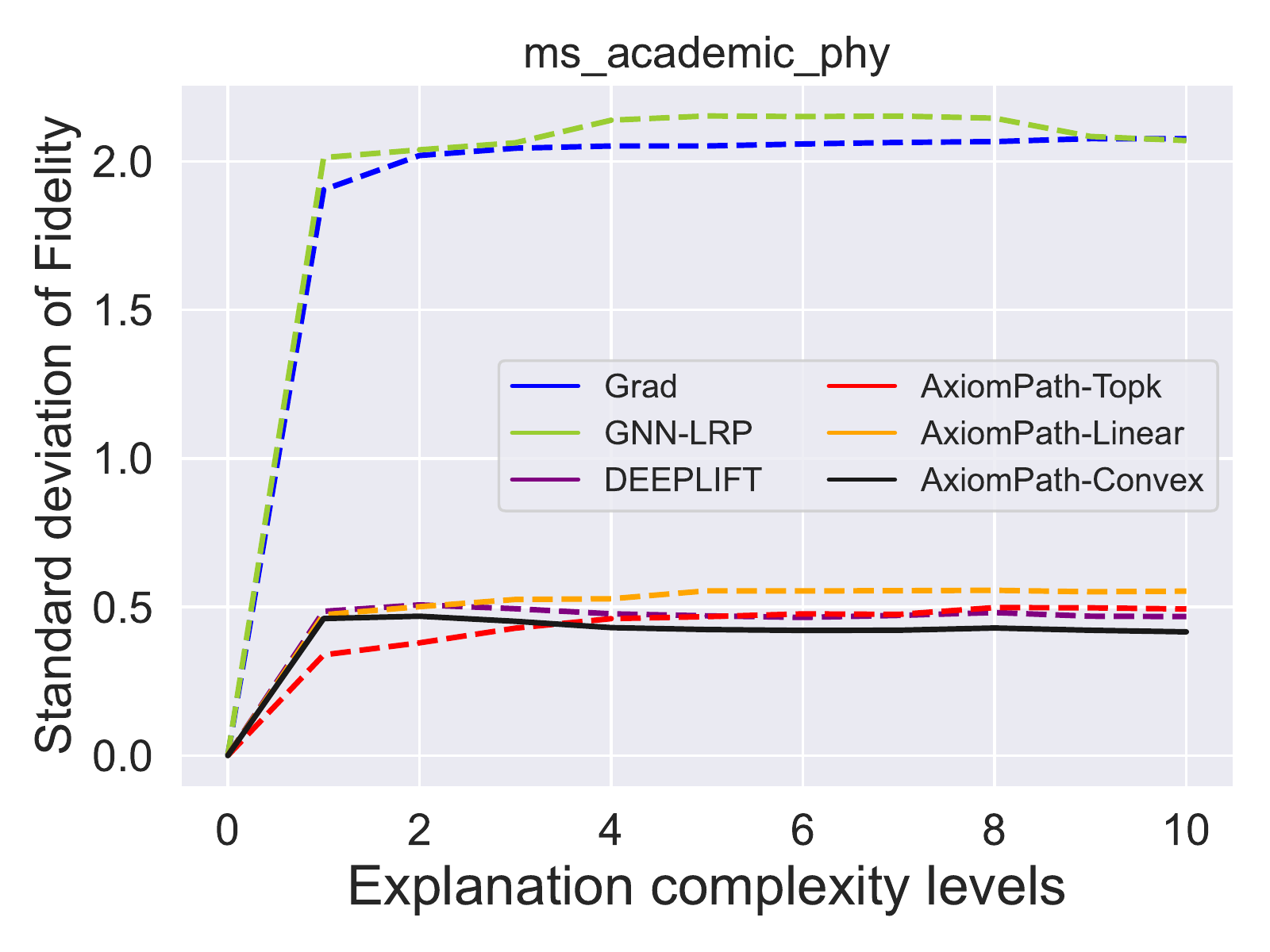}
  }
  \caption{\small Standard deviation of \textbf{Fidelity}$_\textnormal{KL}^{-}$ over all datasets when the number of layers of the fixed GNN model is three.
  }
\label{fig:overall_layer3_std}
\end{figure}

When the number of GNN model layers is three, AxiomPath-Convex can still perform better than other baselines. AxiomPath-Convex has the smallest fidelity over all levels of explanation complexities and over all datasets. AxiomPath-Convex has the smallest standard deviation over all levels of explanation complexities on five datasets
(Cora, Citeseer, Amazon-C, Amazon-P and Coauthor-Computer). In Figure~\ref{fig:overall_layer3_kl} and Figure~\ref{fig:overall_layer3_std}, we demonstrate the effectiveness of the salient path selection of AxiomPath-Convex.

\subsubsection{Running time overhead of convex optimization.} 
\begin{figure}[htb]
  \subfloat{\includegraphics[width = 0.25\textwidth]{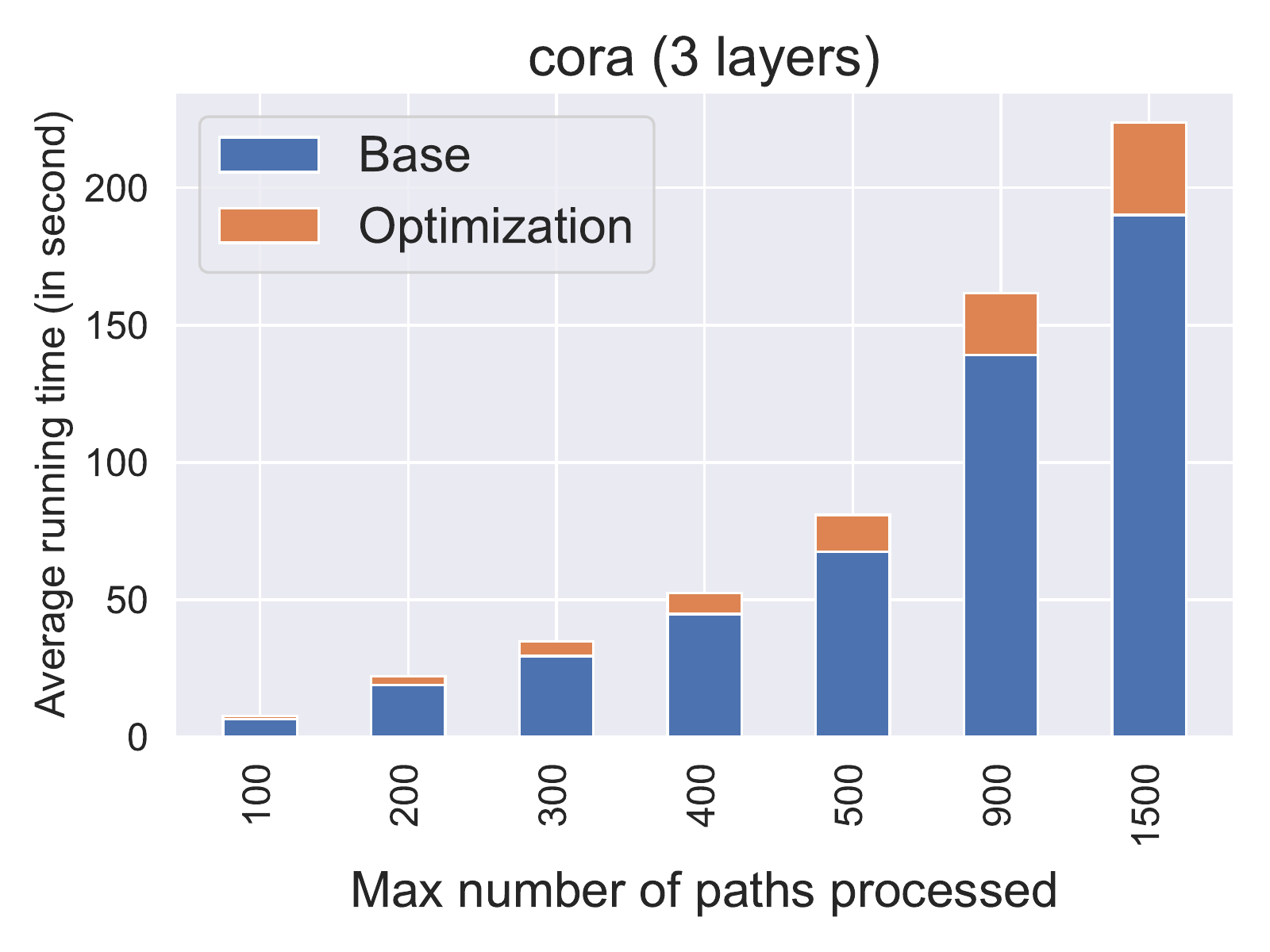}
  }
  \subfloat{\includegraphics[width = 0.25\textwidth]{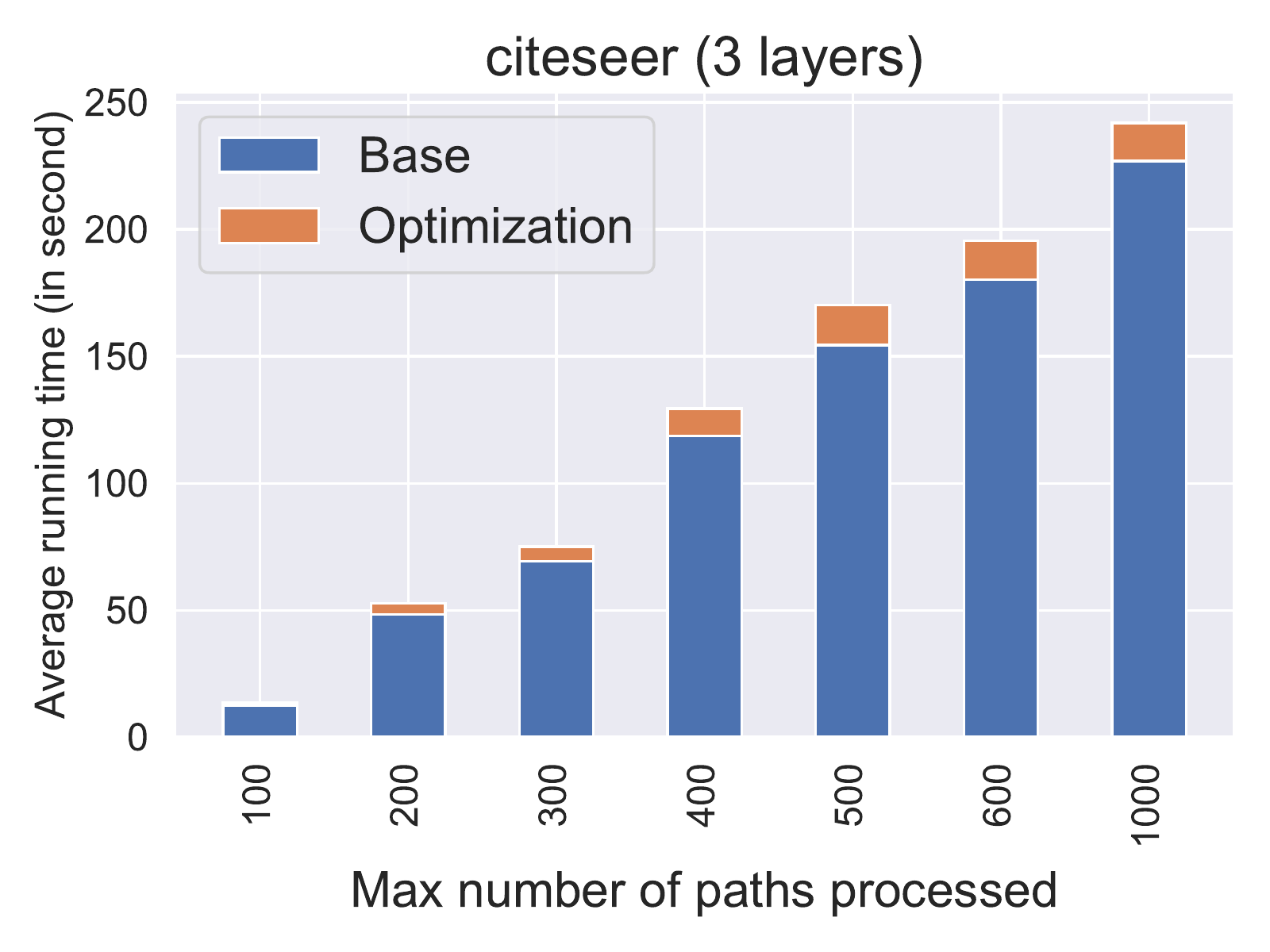}
  }\\
  \subfloat{\includegraphics[width = 0.25\textwidth]{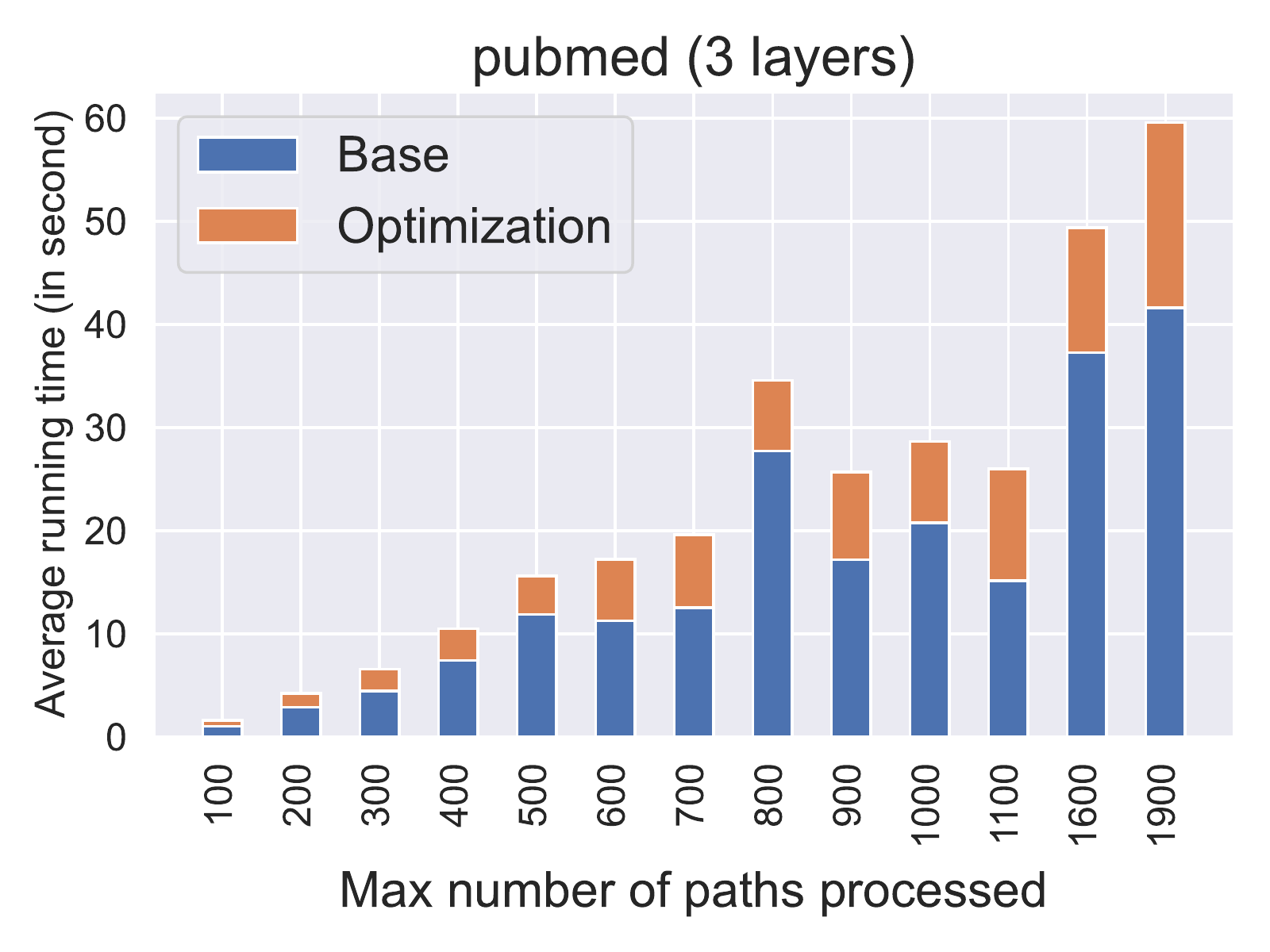}
  }
  \subfloat{\includegraphics[width = 0.25\textwidth]{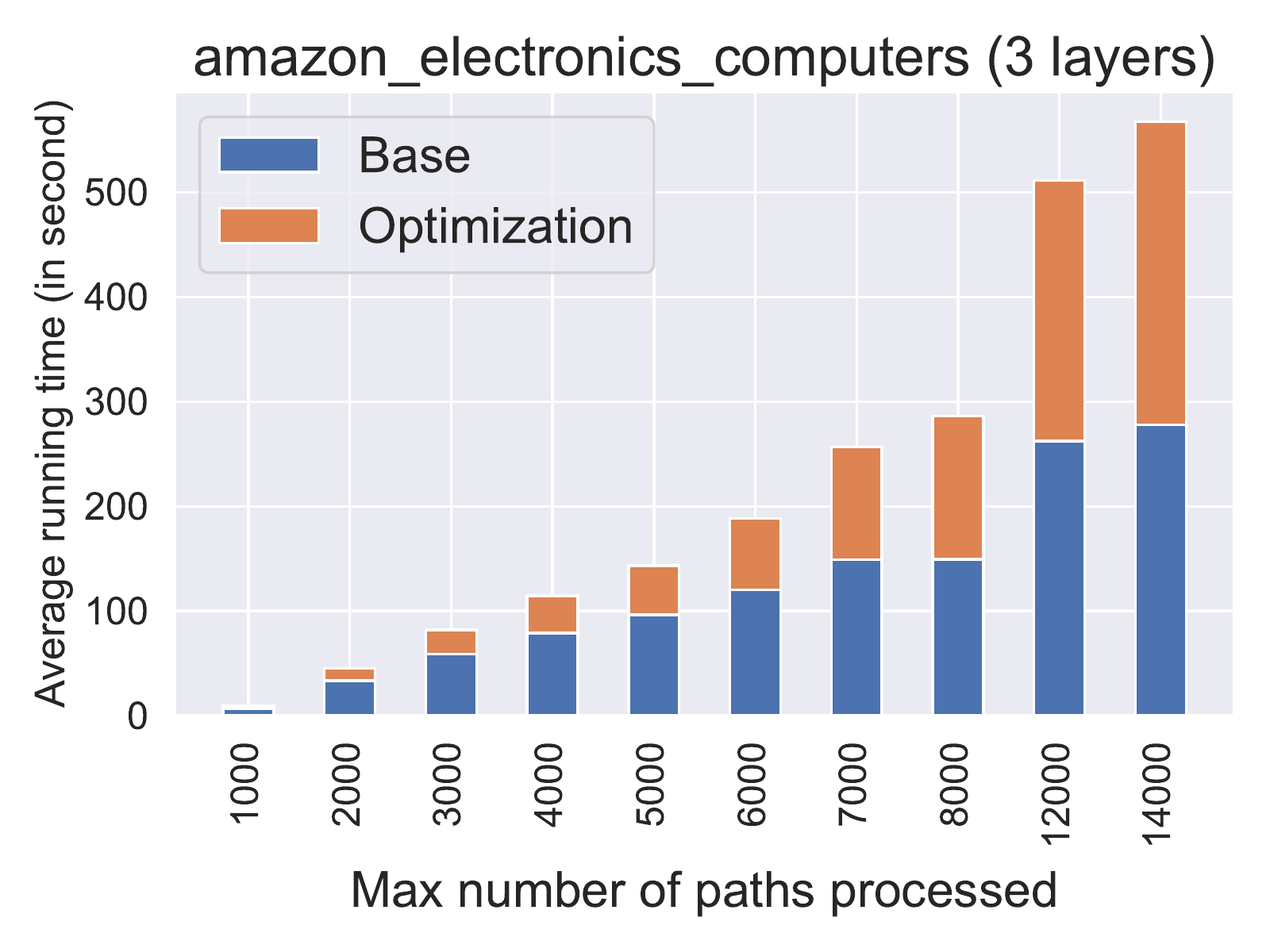}
  }\\
  \subfloat{\includegraphics[width = 0.25\textwidth]{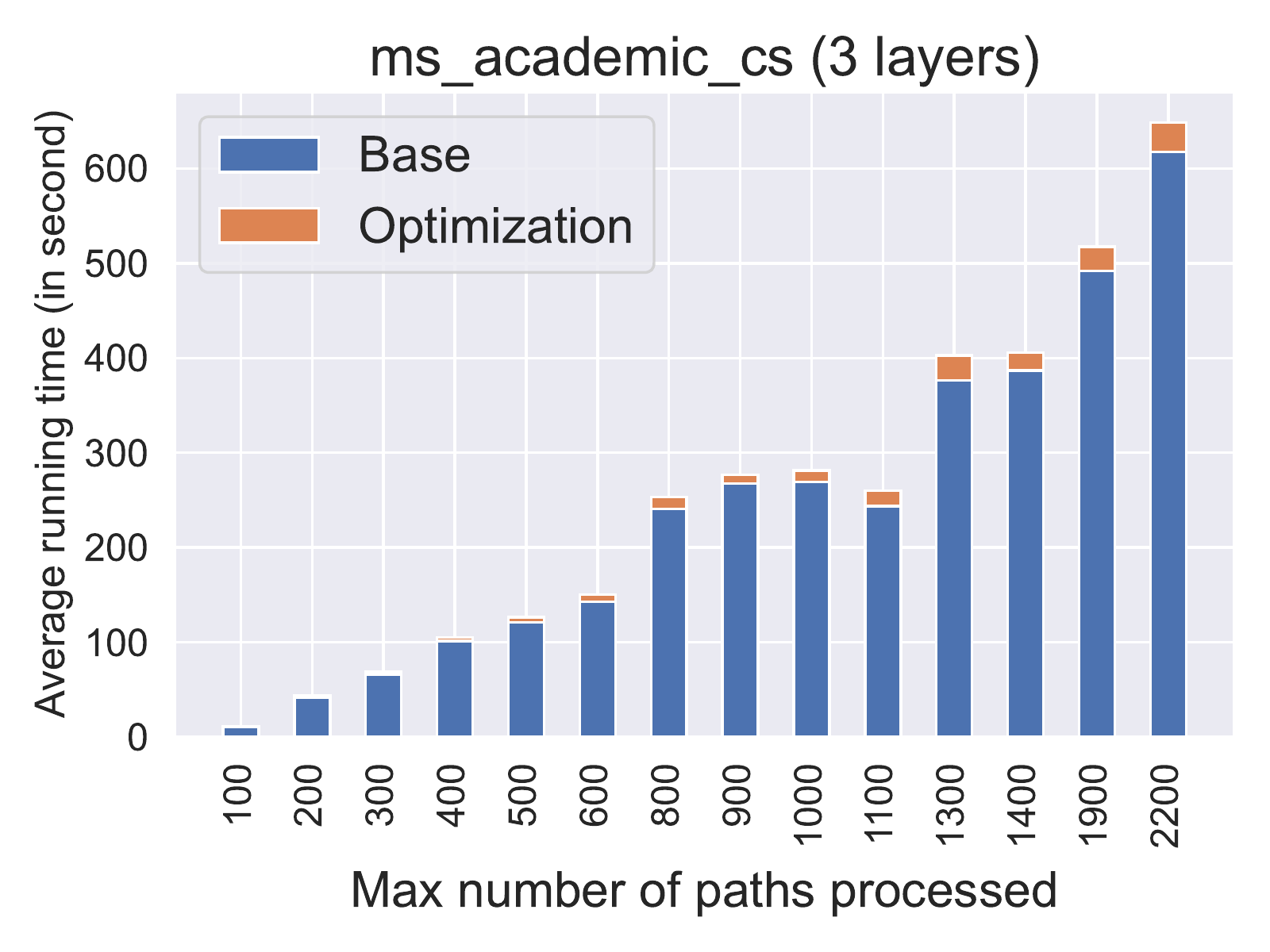}
  }
  \caption{\small Running time of optimization
  }
\label{fig:overall_layer3_time}
\end{figure}

In the main texts, we plot the base running time of searching added path and attribution vs. the running time of optimization afterwards on the  Amazon-Photo and  Coauthor-Physics datasets. 
In Figure~\ref{fig:overall_layer3_time}, we plot the running time on the other five datasets.

We can see that in most cases when the number
of new paths is not extremely large, the optimization step
takes a reasonable amount of time compared to the base running time.
\end{document}